\documentclass{article} 
\usepackage{iclr2027_conference,times}

\usepackage{amsmath,amsfonts,bm}

\def\eqref#1{equation~\ref{#1}}

\def\1{\bm{1}}

\DeclareMathAlphabet{\mathsfit}{\encodingdefault}{\sfdefault}{m}{sl}
\SetMathAlphabet{\mathsfit}{bold}{\encodingdefault}{\sfdefault}{bx}{n}

\usepackage{hyperref}
\usepackage{url}
\usepackage{natbib}
\usepackage{algorithm}
\usepackage{algpseudocode}
\usepackage{cleveref}
\usepackage{booktabs}
\usepackage{graphicx}
\usepackage[most]{tcolorbox}
\usepackage[T1]{fontenc}
\newtcblisting[auto counter]{promptbox}[2][]{enhanced, breakable, colback=gray!5, colframe=black!60, boxrule=0.4pt, arc=1pt, left=2pt, right=2pt, top=2pt, bottom=2pt, fonttitle=\small\bfseries, title={Box~\thetcbcounter: #2}, listing only, listing options={basicstyle=\ttfamily\scriptsize, breaklines=true, breakindent=1em, columns=fullflexible, keepspaces=true, showstringspaces=false}, #1}
\crefname{tcb@cnt@promptbox}{Box}{Boxes}
\Crefname{tcb@cnt@promptbox}{Box}{Boxes}
\title{Constrained Command-Conditioned Reinforcement Learning with Bandit Strategy Selection in Real-Time Strategy Games}

\author{Nick Leenders\thanks{Corresponding author: \texttt{nick.leenders@nlr.nl}}
\\
	NLR Royal Netherlands Aerospace Centre\\
	Amsterdam, The Netherlands \\
	\And 
	Joost van Oijen \\
	NLR Royal Netherlands Aerospace Centre\\
	Amsterdam, The Netherlands \\
	\AND
	Roy Lindelauf \\
    Data Science Center of Excellence \\
    Faculty of Military Sciences, Breda, NL
	\And
	Boris Čule \\
    Departement Intelligent Systems \\
    Tilburg University, Tilburg, NL
}

\iclrfinalcopy 
\begin{document}

\maketitle
\lhead{Preprint. Under review.} 
 
\begin{abstract}
Deep reinforcement learning agents reach strong performance in real-time strategy games but can be brittle against opponents outside their training distribution. Separating strategic command selection from learned unit control allows different strategies to be selected for different opponents while reusing the same execution policy. This requires an executor that can follow different commands and measurable criteria for assessing whether it does so.
We introduce a constrained command-conditioned Proximal Policy Optimization (PPO) policy, the executor, for MicroRTS, a real-time strategy environment. Discrete commands specify strategic objectives and behavioral requirements for economy, army composition, military posture, and worker policy over multiple environment steps; the executor determines the unit-level actions used to fulfill them.
A Thompson-sampling bandit acts as the strategist, selecting command tuples from an estimate of the opponent's strategy built from in-game observations rather than opponent identity. In a controlled comparison with a flat PPO baseline trained with the same architecture, budget, curriculum and self-play league, the strategist–executor system wins significantly more often against three of the four strongest opponents on a training map, including the two strongest held-out ones (0.55 → 0.97 and 0.01 → 0.34), with no significant difference against the others. 
\end{abstract}

\section{Introduction}
Real-time strategy (RTS) games have been a challenging benchmark for showcasing advances in Reinforcement Learning (RL), with exemplary applications such as  OpenAI Five \citep{openaiDota2Large2019} and DeepMind's AlphaStar \citep{vinyalsGrandmasterLevelStarCraft2019}. Although these agents often reach superhuman performance, they still have limitations in several areas. First, while some post-hoc analysis of learned strategies is possible \citep{openai_openai_2018}, it remains difficult to understand why these agents make particular decisions or to directly steer their behavior. Secondly, these agents show strategic brittleness: they become optimized against recent training opponents but fail against strategies they have not encountered recently. This is typically addressed by maintaining a diverse population of agents during self-play training \citep{vinyals_grandmaster_2019}, which makes robustness a property of the training distribution rather than something the agent can adapt at play time.

Separating strategic choice from unit control is the logical answer to both problems, but existing separations each sacrifice something. Hierarchical RL \citep{dayanFeudalReinforcementLearning,bugeja_n-layered_2019} makes the goal explicit, yet rewards following it with a bonus that, in an adversarial game, trades off directly against winning, and gives no criterion for whether it was followed at all. Portfolio-based search and programmatic strategies \citep{churchillPortfolioGreedySearch2013,marinoProgrammaticStrategiesRealTime2021} provide an interpretable and verifiable choice space, but rely on scripted execution that cannot match learned policies in unit-level control.

We introduce a framework for strategic control built around three components: a fixed, human-readable command space with explicit, state-measurable success criteria; an executor trained across commands to maximize game performance subject to a separate success-rate constraint for each command; and a strategist that adjusts command selection in response to observed opponent behavior without retraining the executor. We evaluate the framework in microRTS \citep{huang_gym-rts_2021}, an RTS environment used in the annual IEEE Conference on Games competitions. We highlight the following main contributions:

\begin{itemize}

\item A command system with measurable outcomes. The strategist selects a four-slot command tuple covering economy, unit composition, military posture, and worker policy, with 16 commands across the four slots. Each command specifies when it may be issued and what the executor must achieve within an execution window, such as producing the requested unit composition. Game-state measurements determine whether the command succeeds, fails, or times out, allowing us to measure how reliably the executor follows each command and, by holding commands fixed, which strategies win against which opponents.
\item A constrained command-conditioned executor. A RL policy trained to maximize game performance subject to a separate success-rate constraint for each command. Per-command Lagrange multipliers adapt to measured adherence shortfalls and weight normalized adherence advantages, with a heuristic guard for persistently stalled constraints.
\item An observation-based bandit strategist. A bandit strategist learns how well command tuples perform against different opponent behaviors, sharing information across combinations through their individual commands. During training, it explores promising and uncertain combinations using Thompson sampling. During evaluation, the frozen strategist selects command tuples from its learned estimates and the opponent behavior suggested by current observations, without opponent identity or retraining the executor.
\item A controlled comparison with flat PPO. We compare the complete strategist--executor system against a flat PPO baseline using the same base architecture, environment-step budget, curriculum, and self-play league. Our strategist--executor achieves significantly higher win rates against three of the four strongest opponents, two of them unseen during training, and matches flat PPO against the weaker ones. 

\end{itemize}
\section{Related Work}
We first review RTS game AI, mostly developed in StarCraft and MicroRTS, and then more general work on goal-conditioned and hierarchical RL and on strategy selection against unknown opponents.

\paragraph{Scripted portfolios and action abstraction.}
Pure RL approaches such as AlphaStar \citep{vinyals_grandmaster_2019} and OpenAI Five \citep{openai_dota_2019} are steered only by retraining; their strategy lives implicitly in the weights. Scripted portfolios and action abstraction make strategic decisions explicit instead. Portfolio Greedy Search assigns each unit a script from a fixed, hand-written portfolio \citep{churchillPortfolioGreedySearch2013}; related methods select one script per unit type \citep{lelisStratifiedStrategySelection2017}, expose script choice points to look-ahead search \citep{barrigaPuppetSearchEnhancing2015}, let only a selected subset of units leave script-generated actions \citep{moraesAsymmetricActionAbstractions2018}, or pair a learned strategic layer with scripted execution \citep{barrigaCombiningStrategicLearning2021}. Programmatic-strategy methods synthesize the complete policy as a program in a domain-specific language \citep{marinoProgrammaticStrategiesRealTime2021}; InnateCoder guides this search with foundation-model generated programs and is competitive with past MicroRTS competition winners \citep{moraesInnateCoderLearningProgrammatic2025}. In all of these, low-level behavior is specified by scripts or a program space rather than learned. We retain the idea of a small, fixed strategic vocabulary, but each vocabulary item specifies a goal with a state-measurable success criterion, while a learned executor determines how to achieve it.

\paragraph{Goal-conditioned and hierarchical RL.}
FeUdal Networks \citep{vezhnevetsFeUdalNetworksHierarchical} represent goals as directions in a learned latent space and train the low-level policy using both task reward and an intrinsic reward for directional alignment; \citet{bugeja_n-layered_2019} extend this approach to an N-Layered FeUdal Network evaluated on StarCraft II minigames. These methods measure alignment with learned goals, but do not associate each goal with a predefined, human-readable success criterion.

Explicit goals and interpretable abstractions also appear in prior work. h-DQN \citep{kulkarniHierarchicalDeepReinforcement} trains a goal-conditioned controller to maximize intrinsic reward for achieving explicit goals, while its high-level policy optimizes task reward. Policy sketches \citep{andreasModularMultitaskReinforcement} learn a modular sub-policy for each named subtask from task-level reward alone. \citet{jiangLanguageAbstractionHierarchical} use language instructions as the interface to a reusable controller trained with instruction-achievement rewards and hindsight relabeling.

A closely related approach in RTS games is Grounded RL \citep{xuGroundedReinforcementLearning}, which uses constrained RL to learn to win while following human commands. It measures command adherence indirectly by bounding the behavior-cloning loss on human demonstrations. Our approach instead measures adherence through explicit success criteria defined in terms of the game state. Each command in a fixed, human-readable vocabulary has its own success-rate target, with a separate Lagrange multiplier adapted to measured shortfalls. The executor learns to meet these targets through interaction, without human demonstrations.



\paragraph{Strategy selection and opponent adaptation.}
Selecting among reusable behaviors provides a way to respond to an unfamiliar opponent without learning an entire policy from scratch. Bayesian Policy Reuse \citep{rosmanBayesianPolicyReuse2016} selects from a learned policy library using a belief over task types inferred from observed signals and a model of policy performance. Deep BPR+ \citep{zhengDeepBayesianPolicy} extends this approach to non-stationary opponents, identifying response policies from rewards and observed opponent behavior. In RTS games, \citet{gehring2018high} learn to switch among rule-based build orders in StarCraft: Brood War from partial observations, using hidden opponent unit counts as an auxiliary prediction target during training. Bandit methods also operate at a lower decision level: \citet{ontanonCombinatorialMultiArmedBandit2013} use combinatorial bandits to select combinations of unit actions within Monte Carlo tree search. Our strategist selects compositional command tuples that a single shared policy executes, rather than choosing from separately trained policies or predefined build orders. Payoff evidence is shared across tuples with common commands.

\section{Method}
\label{sec:method}
\subsection{Strategist--Executor Interface}
\begin{figure}[h]
    \centering
    \includegraphics[width=0.99\linewidth]{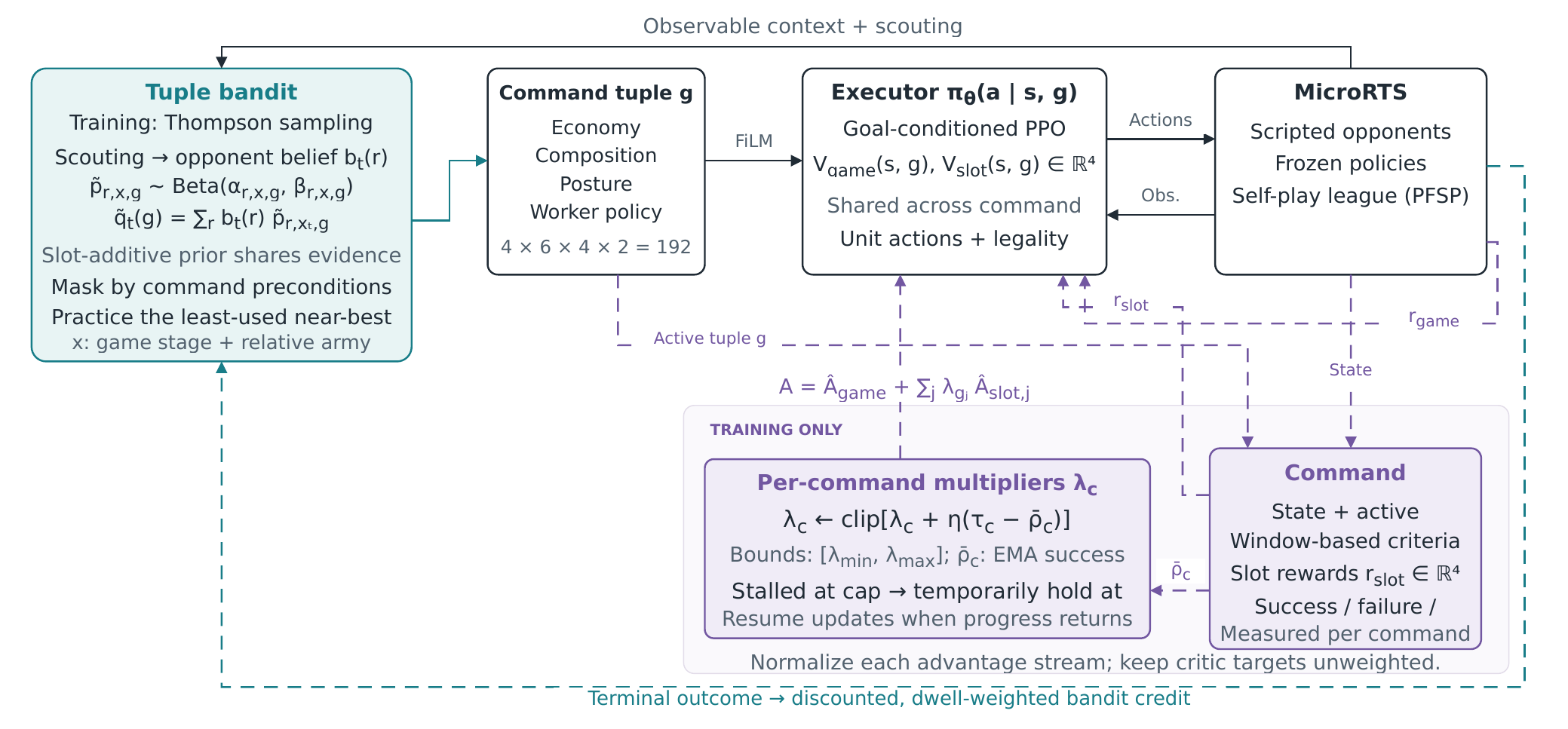}
  \caption{Overview of the framework. The tuple bandit (strategist) supplies commands to a shared executor during training and evaluation. Dashed arrows show training feedback: terminal game outcomes update the bandit, while measured command outcomes adapt the multipliers used in PPO.}
    \label{fig:overview}
\end{figure}
The strategist selects high-level commands that specify strategic objectives and behavioral requirements over multiple environment steps, while the executor determines the unit-level actions used to fulfill them. We use “command” for an instruction in this interface, with fulfillment defined by a state-measurable success criterion.
\Cref{fig:overview} summarizes the framework. The strategist issues a four-slot command tuple, and the goal-conditioned executor translates it into unit-level actions in MicroRTS, where workers gather resources and build bases, which train workers, and barracks, which train Light, Heavy or Ranged military units (Appendix~\ref{sec:microrts_env}). We write $s_t$ for the information available to the executor at game tick (environment step) $t$, $\mathbf{g}_t$ for the active command tuple, and $\pi_\theta(a_t\mid s_t,\mathbf{g}_t)$ for the shared execution policy.

\subsection{Command system}
The executor is conditioned on a four-slot command tuple representing economy,
unit composition, military posture, and worker policy:
\begin{equation}
    \mathbf{g}_t =
\left(
g_t^{\mathrm{eco}},
g_t^{\mathrm{comp}},
g_t^{\mathrm{post}},
g_t^{\mathrm{worker}}
\right)
\in
\mathcal{C}_{\mathrm{eco}}
\times
\mathcal{C}_{\mathrm{comp}}
\times
\mathcal{C}_{\mathrm{post}}
\times
\mathcal{C}_{\mathrm{worker}}.
\end{equation}

The four command vocabularies contain 4, 6, 4, and 2 commands, respectively, yielding a tuple space $\mathcal{G}$ of 192 combinations. Each command has a criterion evaluated over a temporal window, with a deadline and a terminal outcome in $\{\mathrm{success},\mathrm{timeout},\mathrm{fail}\}$ (\Cref{tab:command_outcomes}). Installing a different command opens a new window for that slot. Reissuing an unchanged active command preserves its window, so repeated commands cannot keep restarting the deadline.

Issue preconditions determine which commands may be selected in the current state. Separate evaluation rules keep the success rate meaningful: a command should receive credit only when there was a valid opportunity to demonstrate the requested behavior. Windows with no valid observations are therefore not automatically counted as successes. If a new command interrupts a window before a terminal event occurs, that window is excluded from the success-rate calculation because its outcome is unresolved. The command vocabulary is shown in \Cref{tab:rl_commands} and issue preconditions are listed in Appendix~\ref{app:issue_preconditions}.

\begin{table}[h]
\centering
\caption{Command vocabulary of the low-level RL executor. One command is active in each slot.}
\label{tab:rl_commands}
\begin{tabular}{llll}
\toprule
\textbf{Economy}
& \textbf{Composition}
& \textbf{Military posture}
& \textbf{Worker policy} \\
\midrule
\texttt{STABILIZE}
& \texttt{LIGHT}
& \texttt{OFFENSE}
& \texttt{ECON\_ONLY} \\

\texttt{GROW}
& \texttt{HEAVY}
& \texttt{DEFENSE}
& \texttt{HARASS} \\

\texttt{HOLD}
& \texttt{RANGED}
& \texttt{SPLIT} \\

\texttt{DOUBLE\_BARRACKS}
& \texttt{LIGHT\_HEAVY}
& \texttt{FLANK}
& \\

&
\texttt{LIGHT\_RANGED}
&
& \\

&
\texttt{HEAVY\_RANGED}
&
& \\
\bottomrule
\end{tabular}
\end{table}

\subsection{Executor}
The executor is a GridNet \citep{hanGridWiseControlMultiAgent} policy trained with PPO \citep{schulmanProximalPolicyOptimization2017}. It produces per-cell multi-discrete actions with invalid-action masking \citep{huangInvalidActionMasking2020}. A convolutional encoder and residual blocks \citep{heDeepResidualLearning2016} produce spatial features; a decoder upsamples these features with a command-gated skip connection to produce per-cell logits. The same policy serves all command tuples. Architecture details are given in Appendix~\ref{sec:architecture}.

The command tuple conditions the network through Feature-wise Linear Modulation (FiLM) \citep{perezFiLMVisualReasoning2017}. Let $\mathbf{f}_t\in\mathbb{R}^{C\times H\times W}$ denote the spatial features produced by the convolutional encoder from $s_t$. The slot-preserving concatenation of command embeddings is mapped to channel-wise FiLM parameters:
\begin{equation}
[\boldsymbol{\gamma}_t,\boldsymbol{\beta}_t]
=
F_{\psi}\!\left(\mathbf{e}(\mathbf{g}_t)\right),
\qquad
\widetilde{\mathbf{f}}_t
=
(1+\boldsymbol{\gamma}_t)\odot\mathbf{f}_t
+
\boldsymbol{\beta}_t.
\end{equation}
Here, $\odot$ denotes channel-wise multiplication, applied at each of the $H\times W$ spatial positions, and $\widetilde{\mathbf{f}}_t$ denotes the command-conditioned features. The actor and both value functions use command-conditioned features, so the same observed state can lead to different actions and value estimates under different commands. The game reward $r_t^{\mathrm{game}}$, which combines the terminal game outcome with in-game progress signals such as resource collection, construction, attacking, and military-unit production, is kept separate from the four slot-adherence rewards $r_{t,j}^{\mathrm{slot}}$, each of which combines progress-based shaping within the command's window with a terminal payment when the window resolves. Reward definitions are collected in Appendix~\ref{app:reward_structure}.

\paragraph{Command-success constraints.}
Let $\mathcal{C}$ be the set of individual commands, and let $\rho_c(\theta)$ be the probability of success among evaluable windows for command $c$ that produce a terminal outcome, under the current training distribution of opponents and commands. The intended executor objective is
\begin{equation}
\max_\theta J_{\mathrm{game}}(\theta)
\quad\text{subject to}\quad
\rho_c(\theta)\geq\tau_c\quad\forall c\in\mathcal{C},
\qquad
J_{\mathrm{game}}(\theta)=\mathbb{E}\!\left[\sum_t\gamma^t r_t^{\mathrm{game}}\right],
\label{eq:command_constraints}
\end{equation}
where $\tau_c$ is the command's success-rate target. Introducing a nonnegative multiplier for each constraint gives the Lagrangian \citep{altman2021constrained}
\begin{equation}
\mathcal{L}(\theta,\boldsymbol{\lambda})
=J_{\mathrm{game}}(\theta)
+\sum_{c\in\mathcal{C}}\lambda_c\bigl(\rho_c(\theta)-\tau_c\bigr),
\qquad \lambda_c\geq 0.
\label{eq:command_lagrangian}
\end{equation}

\paragraph{PPO with separate advantage streams.}
We use the shaped slot rewards to guide policy learning and terminal command outcomes to adapt the multipliers, following the distinction between a guiding reward and a measured constraint in RCPO \citep{tesslerRewardConstrainedPolicy2018}. A scalar game critic $V_{\mathrm{game}}(s_t,\mathbf{g}_t)$ and a four-output adherence critic $V_{\mathrm{slot}}(s_t,\mathbf{g}_t)\in\mathbb{R}^4$ estimate their respective returns. Generalized Advantage Estimation \citep{schulman_high-dimensional_2016} is applied independently to each reward stream, yielding the PPO advantage
\begin{equation}
A_t^{\lambda}
=
\mathcal{N}\!\left(\widehat{A}_t^{\mathrm{game}}\right)
+
\sum_{j=1}^{4}\lambda_{g_{t,j}}\,
\mathcal{N}\!\left(\widehat{A}_{t,j}^{\mathrm{slot}}\right),
\label{eq:combined_advantage}
\end{equation}
where $\mathcal{N}$ normalizes each stream using its mean and standard deviation over the learner batch and $g_{t,j}$ denotes the command active in slot $j$. Normalizing the streams separately prevents differences in reward scale from determining their relative influence, allowing the multipliers to control the weight given to command adherence. PPO uses $A_t^{\lambda}$ in its clipped policy objective. The critics are trained on the raw, unweighted reward streams, so their targets do not shift when a multiplier changes; the multipliers enter only when advantages are combined. Because the actor is trained on shaped per-step rewards with normalized advantages, rather than on the terminal success rates in \Cref{eq:command_constraints}, this update approximates the Lagrangian objective rather than following its exact gradient.

\paragraph{Multiplier updates.}
At PPO update $k$, let $S_{c,k}$, $F_{c,k}$, and $T_{c,k}$ count terminal successes, failures, and timeouts for command $c$ in the collected rollout. We smooth the measured success rate and update the multiplier by
\begin{equation}
\widehat{\rho}_{c,k}=\frac{S_{c,k}}{S_{c,k}+F_{c,k}+T_{c,k}}, \qquad \bar\rho_{c,k}=\nu\bar\rho_{c,k-1}+(1-\nu)\widehat{\rho}_{c,k}
\label{eq:command_rate}\\
\end{equation}
\begin{equation}
\lambda_{c,k+1}=\operatorname{clip}_{[\lambda_{\min},\lambda_{\max}]}
\left[\lambda_{c,k}+\eta_\lambda\bigl(\tau_c-\bar\rho_{c,k}\bigr)\right]
\label{eq:dual_update}
\end{equation}
Here $\nu$ is a smoothing coefficient and $\eta_\lambda$ the multiplier step size. A rollout may contain only a few outcomes for a given command, making its measured success rate noisy. Smoothing reduces fluctuations between rollouts, and updates with too few terminal events leave both the rate estimate and multiplier unchanged.  A rate below target increases the weight of that command's adherence advantage; a rate above target reduces it. This is the violation-driven dual update associated with \Cref{eq:command_lagrangian}, with finite lower and upper bounds for stability.

During training, the executor may not yet have learned to carry out some commands reliably. Giving these commands a large weight can divert learning from game performance without improving command execution. We therefore use a heuristic guard: if a command's success rate remains below target and stops improving while its multiplier stays near the upper bound, we temporarily reduce that multiplier to its minimum value. Normal multiplier updates resume once the success rate improves again. The guard does not prove that a target is infeasible or guarantee that it will be met. Settings are given in Appendix~\ref{app:multiplier_settings}.

\subsection{Strategist}
The strategist maintains a payoff table indexed by opponent class $r$ (one class per training opponent), observable context $x$, and command tuple $\mathbf{g}$. The context summarizes game stage and relative military strength. Each cell holds discounted win and non-win counts, $W_{r,x,\mathbf{g}}$ and $L_{r,x,\mathbf{g}}$. The strategist revisits the active tuple only at certain decision points: at the start of the game, periodically during play (usually about every 150 ticks), and when newly scouted evidence persists (Appendix~\ref{app:tuple_bandit}).

\paragraph{Scouting-based opponent belief.}
At each decision point, an evidence key $e_t$ summarizes what has been observed of the opponent so far: enemy structures, military unit types, worker aggression, worker counts, and elapsed game time. It is built from observations only, without the opponent's identity. During training, a recognizer counts how often each key occurs with each opponent class, $n(e,r)$, using the known class as the label. Smoothing these counts with a pseudocount, which has the form of a Dirichlet-categorical posterior mean \citep[Section~2.2]{bishopPatternRecognition2006}, gives
\begin{equation}
b_t^{\mathrm{raw}}(r)
=\frac{n(e_t,r)+a_0}
{\sum_{r'\in\mathcal{R}}n(e_t,r')+a_0|\mathcal{R}|},
\qquad a_0>0,
\label{eq:scouting_belief}
\end{equation}
where $\mathcal{R}$ is the set of known training classes and $a_0$ a symmetric pseudocount, so an unseen evidence key gives a uniform belief. Dropping negligible classes and renormalizing gives the selection weights $b_t(r)$ (Appendix~\ref{app:tuple_bandit}). At evaluation the counts stay fixed, but new evidence changes the queried key, so an unseen opponent is represented as a mixture of known classes.

\paragraph{Payoff estimates and Thompson sampling.}
For each class, context, and tuple, we form Beta parameters
\begin{equation}
\alpha_{r,x,\mathbf{g}}=W_{r,x,\mathbf{g}}+\kappa m_{r,x,\mathbf{g}},
\qquad
\beta_{r,x,\mathbf{g}}=L_{r,x,\mathbf{g}}+\kappa(1-m_{r,x,\mathbf{g}}),
\label{eq:bandit_beta}
\end{equation}
where $\kappa$ is the prior strength. The prior mean $m_{r,x,\mathbf{g}}$ combines the tuple's per-slot outcome statistics additively in log-odds space, relative to the class--context base rate. Thus, an unplayed tuple inherits evidence about its constituent commands; the prior construction is detailed in Appendix~\ref{app:tuple_bandit}.

At each training-time bandit decision, Thompson sampling \citep{thompsonLikelihoodThatOne1933,russoTutorialThompsonSampling2018} draws a payoff estimate for every candidate tuple and retained opponent class, then averages these draws under the scouting belief:
\begin{equation}
\widetilde p_{r,x_t,\mathbf{g}}
\sim\operatorname{Beta}(\alpha_{r,x_t,\mathbf{g}},\beta_{r,x_t,\mathbf{g}}),
\qquad
\widetilde q_t(\mathbf{g})
=\sum_{r\in\mathcal{R}} b_t(r)\widetilde p_{r,x_t,\mathbf{g}}.
\label{eq:thompson_score}
\end{equation} 

Let $\mathcal{F}_t$ denote tuples satisfying command preconditions and, after the opening, differing from the current tuple in at most one slot, and let $N_t(x_t,\mathbf{g})$ count previous bandit selections in the current context. To encourage practice beyond a single estimated best tuple, the proposed tuple is
\begin{equation}
\mathbf{g}_t^*\in\arg\min_{\mathbf{g}\in\mathcal{F}_t}N_t(x_t,\mathbf{g})
\quad\text{subject to}\quad
\widetilde q_t(\mathbf{g})\geq
\max_{\mathbf{g}'\in\mathcal{F}_t}\widetilde q_t(\mathbf{g}')-\delta,
\label{eq:bandit_selection}
\end{equation}
with random tie-breaking. Two rules limit rapid switching: the current tuple must have been active for a minimum time before it can be replaced, and the proposal must improve on the current feasible tuple's score by a margin.

\paragraph{Evaluation.}
The executor, payoff table, and recognizer remain frozen. We replace sampled payoffs with posterior means,
\begin{equation}
q_t(\mathbf{g})
=\sum_{r\in\mathcal{R}} b_t(r)
\frac{\alpha_{r,x_t,\mathbf{g}}}
{\alpha_{r,x_t,\mathbf{g}}+\beta_{r,x_t,\mathbf{g}}},
\label{eq:bandit_mean}
\end{equation}
and propose a maximizing feasible candidate, retaining both switching rules. Adaptation at evaluation therefore comes only from changing context and scouting evidence.

\subsection{Training procedure}
The executor and strategist are trained together from scratch in a single run. Each episode takes commands from either the strategist or a generated schedule (random command tuples switched at preset ticks or game states). Early episodes use schedules only, to cover the command space; the strategist's share then grows, but a fixed share of schedule episodes remains so the executor keeps practicing tuples the strategist does not select. Both kinds of episodes train the executor and supply scouting evidence and game outcomes to the bandit. A game may use several tuples but yields one outcome, a win or a non-win, which we split among the tuples in proportion to how long each one was active. This avoids crediting only the final tuple but does not identify each tuple's causal contribution. Affected table rows are discounted before each update, so older results lose weight as the executor changes (Appendix~\ref{app:tuple_bandit}).

Opponents are drawn from a scripted-bot curriculum and from a self-play league of a frozen policy and periodic executor snapshots; prioritized fictitious self-play \citep{vinyals_grandmaster_2019} selects within the pool based on recent results. The comparison with the flat PPO uses the same curriculum, pool, and interaction budget. Appendix~\ref{app:training} gives the training settings.

\section{Results}

\subsection{Command adherence and controllability}
\label{sec:controllability}
A success-rate constraint measures whether a command's criterion is met on the states in which the command is issued. It does not measure whether issuing the command changes what the executor does. We therefore separate two questions: whether the constraints were satisfied during training, and whether each command has a causal effect on behavior under an intervention test.

\paragraph{Constraints are satisfied.} At the end of training all sixteen smoothed success rates $\bar\rho_c$ lie between $0.75$ and $1.00$ against a target of $0.8$, no constraint is flagged as stalled, and the multipliers have settled between the lower bound and $0.83$ (\Cref{tab:final_duals}). By the criterion the executor was trained on, every command is at or within $0.05$ of its target.

\paragraph{Intervention test.} We freeze the executor and remove the strategist. All games open with the same tuple; once the executor reaches a mid-game state in which every issue precondition holds, we install either a reference tuple (\texttt{HOLD}, \texttt{LIGHT}, \texttt{DEFENSE}, \texttt{ECON\_ONLY}) or a tuple that differs from it in one slot, and compare each game with the reference game of the same policy seed. Against coacAI the paired games are identical up to the intervention. We measure physical quantities rather than the training criterion (\Cref{tab:controllability}), with army position taken on the axis from the own base (0) to the enemy base (1). The test covers 134 games against three opponents; Appendix~\ref{app:intervention} gives the protocol.

\begin{table}[t]
\centering
\small
\setlength{\tabcolsep}{3.5pt}
\caption{Constraint satisfaction versus controllability. \emph{Train rate} is the smoothed terminal success rate $\bar\rho_c$ at the end of training (target $0.8$). \emph{Response} is the paired difference between the intervention and the same-seed reference game, pooled over three opponents, with a 95\% bootstrap interval over pairs. Metrics use the first 300 ticks after the command, or the whole window when a game ended earlier.}
\label{tab:controllability}
\begin{tabular}{llcccl}
\toprule
Command & Physical metric & Train rate & Reference & Intervention & Response [95\% interval] \\
\midrule
\texttt{GROW}    & new workers            & 0.82 & 0.91 & 2.23 & $+1.32$ [$+1.00$, $+1.68$] \\
\texttt{HEAVY}   & Heavy share of new army & 0.75 & 0.00 & 1.00 & $+1.00$ [$+1.00$, $+1.00$] \\
\texttt{RANGED}  & Ranged share of new army & 0.80 & 0.00 & 1.00 & $+1.00$ [$+1.00$, $+1.00$] \\
\midrule
\texttt{OFFENSE} & army position     & 0.79 & 0.469 & 0.463 & $-0.006$ [$-0.017$, $+0.006$] \\
\texttt{HARASS}  & share workers in enemy half & 0.79 & 0.109 & 0.088 & $-0.021$ [$-0.067$, $+0.023$] \\
\bottomrule
\end{tabular}
\end{table}

\paragraph{Intervention results.} Economy and composition commands change the executor's behavior; posture and worker-policy commands fail to do so (\Cref{tab:controllability}). After \texttt{HEAVY} or \texttt{RANGED}, every new military unit has the requested type, against none under the reference, and \texttt{GROW} adds one to two workers compared with \texttt{HOLD}. These commands can decide the game: from the same state, \texttt{HEAVY} instead of \texttt{LIGHT} turns 8 wins in 8 games against coacAI into 8 losses, and \texttt{RANGED} lowers 7 wins in 7 against obiBotKenobi to 1 in 8. In contrast, \texttt{OFFENSE} does not move the army forward and \texttt{HARASS} does not move workers forward, although both met their constraint in training ($\bar\rho_c=0.79$); both intervals rule out effects larger than $0.07$.

\paragraph{Why the constraints miss this.} A success rate can reach its target while the command has no effect, for three reasons. First, the rate depends on where a command is issued. \texttt{OFFENSE} is issued only once an army exists (Appendix~\ref{app:issue_preconditions}), so it is mostly measured where its criterion is easy to meet. In constructed mid-game states with six units, the executor meets it in 8 of 8 windows but advances equally far under every posture command; in the intervention games against coacAI, which start shortly after the opening, it meets it in 0 of 8. Second, some criteria are met by default behavior. \texttt{DEFENSE} and \texttt{ECON\_ONLY} describe what the executor already does when it plays for the game reward, and the \texttt{HARASS} criterion is met in 15 of 22 windows although the command adds no forward presence. Third, a command without effect earns the same payoff as the default, so the strategist seldom picks it. In the last tenth of training, \texttt{OFFENSE} appears in $3\%$ of plans against coacAI, which leaves the multiplier with few terminal events. In each case the multiplier sees a rate at target, or too few events, and has no reason to increase.

The remedy is therefore a change of measurement, not of objective: the rate in \Cref{eq:dual_update} should be estimated from paired interventions like the one above, which compare a command with another command issued in the same state. We regard this as the main lesson of this study for command-conditioned agents.

\subsection{Evaluation against SOTA opponents}
\label{sec:sota_eval}
We compare the strategist-executor system with a flat PPO policy trained with the same architecture, interaction budget, curriculum and league. Both are frozen, the bandit selects by posterior means (\Cref{eq:bandit_mean}), and every game is played on map A, one of the three training maps. The opponents range from rush scripts to competition winners and two RL agents; six are held out from training (Appendix~\ref{sec:microrts_env}).
Each method is represented by one training run. The statistical tests therefore assess differences between the resulting frozen agents on this map and player seat; they do not account for variability across training runs.

The bandit agent wins at least as often as flat PPO against eight of the twelve opponents (\Cref{tab:model_comparison}). Both agents achieve win rates of at least $0.96$ against the four rush scripts, the weakest opponents in the pool (Appendix~\ref{app:opponent_ranking}), and against \textit{mayari}, \textit{coacAI}, \textit{droplet}, and \textit{aggrobot}. None of these differences is statistically significant.

The largest gains occur against the strongest opponents. Against \textit{agent\_sota}, the bandit agent improves the win rate from $0.61$ to $0.93$. Against the held-out opponents \textit{obiBotKenobi} and \textit{RAISocketAI}, it improves the win rate from $0.55$ to $0.97$ and from $0.01$ to $0.34$, respectively. All three differences are statistically significant. However, \textit{RAISocketAI}, the strongest opponent, still wins most games against both agents. Against \textit{TMA}, the 2024 competition winner and a training opponent, flat PPO wins 100 games and the bandit agent 95; the difference is not significant ($p=0.53$). In their head-to-head evaluation, the bandit agent wins 97 of 100 decisive games. The agents also differ in their unit composition: the bandit agent uses the same Light-unit build against nearly every opponent, whereas flat PPO's build varies with the opponent (Appendix~\ref{app:playstyle}). The gain does not come from adapting commands to the opponent. A fixed tuple selected for each opponent (Static) is not significantly different from the bandit against any opponent, on this map an oracle that picks the best fixed tuple for each opponent gains $0.03$ over the best single tuple in sample and nothing measurable out of sample (Appendix~\ref{app:payoff_matrix}), and apart from a switch to \texttt{RANGED} against coacAI the strategist mostly varies only the economy command (\Cref{sec:strategist_usage}). The difference lies in the executor trained with commands, given a good tuple.

\begin{table}[h]
  \centering
  \small
  \setlength{\tabcolsep}{5pt}
  \renewcommand{\arraystretch}{1.1}
  \caption{Win rates of Flat PPO, the strategist--executor (Bandit) and the executor holding one tuple per opponent, chosen on a separate screen (Static), on map \texttt{16x16/basesWorkers16x16A}, 100 games per cell. Ranks are from \Cref{tab:opponent_ranking} (1 is strongest; rank 10, the training bot \textit{rojo}, is not evaluated). Bold marks the highest observed rate, including ties; $*$ marks significant Flat--Bandit differences ($p<0.05$). Counts, test details and the head-to-head exception ($\dagger$) are in Appendix~\ref{app:model_comparison_details}.}
  \label{tab:model_comparison}
  \begin{tabular*}{\linewidth}{@{\extracolsep{\fill}}rllrrrc@{}}
    \toprule
     & & & \multicolumn{3}{c}{Win rate $W/(W+L)$} & Flat vs Bandit \\
    \cmidrule(lr){4-6}\cmidrule(lr){7-7}
    Rank & Opponent & Split & Flat & Bandit & Static & $p$ \\
    \midrule
     1 & RAISocketAI          & holdout & 0.010 & 0.344 & \textbf{0.470} & $5.9 \times 10^{-10}$\textsuperscript{$*$} \\
     2 & TMA                  & train   & \textbf{1.000} & 0.950 & 0.949 & 0.53 \\
     3 & agent\_sota          & train   & 0.612 & \textbf{0.929} & 0.898 & $6.7 \times 10^{-7}$\textsuperscript{$*$} \\
     4 & obiBotKenobi         & holdout & 0.545 & 0.970 & \textbf{1.000} & $2.0 \times 10^{-12}$\textsuperscript{$*$} \\
     5 & mayari               & train   & \textbf{1.000} & 0.980 & \textbf{1.000} & 1.00 \\
     6 & coacAI               & train   & 0.980 & \textbf{1.000} & \textbf{1.000} & 1.00 \\
     7 & droplet              & holdout & \textbf{0.960} & \textbf{0.960} & 0.950 & 1.00 \\
     8 & aggrobot             & holdout & \textbf{0.990} & 0.970 & 0.950 & 1.00 \\
     9 & workerRushAI         & train   & \textbf{0.980} & 0.970 & 0.970 & 1.00 \\
    11 & lightRushAI          & train   & 0.980 & \textbf{1.000} & \textbf{1.000} & 1.00 \\
    12 & rangedRushAI         & holdout & \textbf{1.000} & \textbf{1.000} & \textbf{1.000} & 1.00 \\
    13 & heavyRushAI          & holdout & \textbf{1.000} & \textbf{1.000} & 0.990 & 1.00 \\
    \midrule
    --- & Head-to-head        &         & 0.030 & 0.970 & \textbf{0.990} & $3 \times 10^{-25}$\textsuperscript{$*\dagger$} \\
    \bottomrule
  \end{tabular*}
\end{table}

\subsection{What the strategist decides during evaluation}
\label{sec:strategist_usage}
\Cref{tab:deployment_usage} reports which commands the strategist issued in the $1\,200$ Bandit games of \Cref{tab:model_comparison}, head-to-head excluded. The posture slot never leaves \texttt{DEFENSE} and the worker slot leaves \texttt{ECON\_ONLY} in 4 games. Every game opens with the same tuple, as no scouting evidence exists at tick zero. The economy order changes at a median tick of 360 to 460: to \texttt{DOUBLE\_BARRACKS} in 94 of 100 games against mayari, and to \texttt{GROW} in at least 95 of 100 games against every other opponent, including all six held-out ones. Against obiBotKenobi, mayari and aggrobot the strategist later moves to \texttt{HOLD} in 51, 36 and 27 games. The composition order depends on the opponent in one case: against coacAI the strategist switches to \texttt{RANGED} in all 100 games, at a median tick of 242, before the economy order changes. The strategist therefore behaves as a selector over the economy and composition commands the executor follows, conditioned on scouting evidence, and holds the remaining two slots at their defaults.

The command interface lets us test how well different strategies work with the same executor. Economy and composition commands change the executor's behavior (\Cref{sec:controllability}), so we can evaluate a strategy by keeping these commands fixed throughout a game. The results (\Cref{tab:payoff_matrix}) show that a Light army achieves the highest win rate, or comes within $0.02$ of it, against every opponent except \textit{coacAI}. Against \textit{coacAI}, \texttt{RANGED} achieves a win rate of $1.00$, while \texttt{HEAVY} achieves $0.00$. However, \texttt{RANGED} wins only $0.11$ of games against \textit{obiBotKenobi} and $0.03$ against \textit{RAISocketAI}.
The strategist consistently selects \texttt{RANGED} against \textit{coacAI}, which agrees with these results.

\begin{table}[t]
\centering
\footnotesize
\setlength{\tabcolsep}{2.5pt}
\caption{Strategist commands in the Bandit games of \Cref{tab:model_comparison}. Every game opens with (\texttt{STABILIZE}, \texttt{LIGHT}, \texttt{DEFENSE}, \texttt{ECON\_ONLY}). \emph{First change}: median tick of the first economy change, which occurs in at least 98 of 100 games per opponent. \emph{Games with a change}: games in which the slot left its opening command; R = \texttt{RANGED}, H = \texttt{HEAVY}, L+H = \texttt{LIGHT\_HEAVY}.}
\label{tab:deployment_usage}
\begin{tabular}{llcccc}
\toprule
 & Most common & First & \multicolumn{3}{c}{Games with a change} \\
\cmidrule(lr){4-6}
Opponent & economy sequence (games) & change & Composition & Posture & Worker \\
\midrule
RAISocketAI  & \texttt{STABILIZE}$\to$\texttt{GROW} (86) & 440 & 18 (H, L+H) & 0 & 0 \\
TMA          & \texttt{STABILIZE}$\to$\texttt{GROW} (97) & 360 & 2 (H) & 0 & 0 \\
agent\_sota  & \texttt{STABILIZE}$\to$\texttt{GROW} (94) & 360 & 9 (H, L+H) & 0 & 0 \\
obiBotKenobi & \texttt{STABILIZE}$\to$\texttt{GROW}$\to$\texttt{HOLD} (50) & 360 & 2 (H) & 0 & 0 \\
mayari       & \texttt{STABILIZE}$\to$\texttt{DOUBLE\_BARRACKS} (46) & 441 & 3 (R) & 0 & 0 \\
coacAI       & \texttt{STABILIZE}$\to$\texttt{GROW} (99) & 460 & 100 (R) & 0 & 0 \\
droplet      & \texttt{STABILIZE}$\to$\texttt{GROW} (100) & 360 & 0 & 0 & 2 \\
aggrobot     & \texttt{STABILIZE}$\to$\texttt{GROW} (70) & 440 & 0 & 0 & 0 \\
workerRushAI & \texttt{STABILIZE}$\to$\texttt{GROW} (100) & 360 & 0 & 0 & 2 \\
lightRushAI  & \texttt{STABILIZE}$\to$\texttt{GROW} (100) & 360 & 0 & 0 & 0 \\
rangedRushAI & \texttt{STABILIZE}$\to$\texttt{GROW} (99) & 360 & 0 & 0 & 0 \\
heavyRushAI  & \texttt{STABILIZE}$\to$\texttt{GROW} (98) & 360 & 24 (R) & 0 & 0 \\
\bottomrule
\end{tabular}
\end{table}

\section{Conclusions}
We introduced a constrained command-conditioned strategist--executor system for real-time strategy games. Measurable command criteria encourage a useful behavioral repertoire, per-command adaptive adherence pressure helps maintain it, and the resulting system wins more often than a flat PPO baseline against three of the four strongest opponents, including the two strongest held-out ones, on a training map, and less often against the 2024 competition winner. This gain, however, stems from command-conditioned training rather than from adaptive behavior switching. Intervention tests show that only economy and composition commands change the executor's behavior, although posture and worker-policy commands also meet their success-rate constraints, and fixed commands chosen with knowledge of the opponent's identity performs on par with the strategist. The result is a versatile agent with steerable strategies. Because these strategies are named commands, which of them wins against which opponent can be measured directly rather than reconstructed from a post-hoc analysis. The agent does not yet switch well between these strategies during play; a next step is to estimate adherence using interventions rather than success rates.

\subsection*{AI use statement}          
We used generative AI tools as coding assistants when implementing methods and experiments, including the evaluation setup, and to provide feedback on methodology and experimental design. We did not use generative AI tools to generate synthetic data, develop theoretical models, formulate or prove mathematical claims, propose hypotheses, clean or reformat datasets, analyze or interpret results, or translate text. We also used generative AI tools to polish the writing of the paper. The authors reviewed and checked all AI-assisted code and AI-suggested edits to the text. We take full responsibility for the content of this paper.

\subsection*{Reproducibility statement}
\Cref{sec:method} and Appendices~\ref{app:command_criteria}--\ref{app:training} specify the command criteria and preconditions, rewards, architecture, strategist, multiplier settings and training schedule. Evaluation protocols (maps, seats, game counts and statistical tests) are given in Appendices~\ref{app:constraints}--\ref{app:payoff_matrix}. All opponents are publicly available (Appendix~\ref{sec:microrts_env}). An anonymized copy of the code is provided as supplementary material at: \url{https://anonymous.4open.science/r/constrained-command-rts/}. Each result comes from a single training run per arm, so reported intervals do not include training-seed variance.

\ificlrfinal
\section*{Funding}
This research is funded by NLR Royal Netherlands Aerospace Centre.
\fi

\bibliography{references}
\bibliographystyle{iclr2027_conference}

\appendix

\section{MicroRTS Environment}
\label{sec:microrts_env}
MicroRTS is a minimalist RTS environment designed specifically as a testbed for RL research. The core gameplay loop requires agents to gather resources via worker units to construct buildings and train military units. Workers can construct two structures: a base, which produces additional workers, and a barracks, which produces distinct military classes: light, ranged, and heavy units. Light units are the cheapest units and therefore are often used in strategies focus on early pressure to the opponents. They are the fastest and are therefore considered a counter to ranged units in most situations. Ranged units are also cheap to produce and are able to attack other units or building from a distance. They have few health points however and easily die once light or especially heavy units get near them. Since they outpace heavy units and can stay away from their reach, they are considered a counter to heavy units. Heavy units are the most expensive units, are quite slow and take long to produce. Light units deal little damage to them so they are considered the counter to light units. Episodes typically begin with a base and at least one worker already in place, however there are a few maps where both sides begin with more bases or barracks. 

MicroRTS presents several classic RTS challenges that the agent must navigate:
\begin{itemize}
\item \textbf{Observation Space:} The environment state is typically represented as a multi-channel spatial grid tensor $\mathcal{S} \in \mathbb{R}^{H \times W \times C}$, where $H$ and $W$ are the map dimensions, and $C$ represents discrete feature planes encoding unit ownership, hit points, unit types, and available resources.
\item \textbf{Action Space:} The action space is highly combinatorial and multi-discrete. Because an agent must control multiple heterogeneous units simultaneously, the branching factor grows exponentially with the number of units on the board.
\item \textbf{Reward Structure:} By default, the environment provides a highly sparse terminal reward. Consequently, intermediate reward shaping is often required to bootstrap early policy learning.
\end{itemize}

To evaluate generalizability, MicroRTS includes a diverse set of maps varying in size, terrain, and resource distribution, requiring distinct macro-strategies to master. Furthermore, the environment features a pool of built-in scripted opponents (such as \textit{coacAI}, \textit{WorkerRush}, and \textit{LightRush}) of varying algorithmic complexity. These baseline bots provide a natural mechanism for curriculum learning, allowing the RL agent to undergo stepped training against increasingly difficult opponents before facing advanced strategies.

\paragraph{Opponent bots and related work.}
The opponents in \Cref{tab:model_comparison} span scripted policies, game-tree search, and deep reinforcement learning. The standard \textit{workerRushAI}, \textit{lightRushAI}, \textit{heavyRushAI}, and \textit{rangedRushAI} baselines use hand-written policies that emphasize attacks with their respective unit types; the Gym-$\mu$RTS benchmark provides the surrounding evaluation framework \citep{huang_gym-rts_2021}. More elaborate scripted opponents include \textit{coacAI} (CoacAI), the 2020 competition winner, and \textit{mayari} (Mayari), the 2021 winner \citep{leCoacAI2020,barvazkravMayari2021,goodfriend_competition_2024}. Both encode resource management, production, and combat decisions in hand-written Java policies. 

The held-out \textit{droplet} bot combines scripts with Monte Carlo tree search. Its implementation uses Guided Na\"ive Sampling to bias search toward actions suggested by scripts \citep{yangDroplet2019}, following the approach of \citet{yangGuidingMonteCarlo2019}. This provides a search-based opponent whose decisions combine predefined behavior with online look-ahead. The other held-out competition bots, \textit{aggrobot} (Aggrobot) and \textit{obiBotKenobi} (ObiBotKenobi), use hand-written control logic \citep{hashimiAggrobot2023,jannisObiBotKenobi2023}. Aggrobot implements explicit unit-group management and movement, production, and attack routines; ObiBotKenobi scores candidate unit actions with role-dependent heuristics and checks conflicts between selected actions. These descriptions follow their released implementations. Both participated in the 2023 competition; ObiBotKenobi placed second on the open maps \citep{goodfriend_competition_2024}.

The opponents trained with RL are \textit{agent\_sota} and \textit{RAISocketAI} \citep{goodfriend_competition_2024}. In our evaluation, \textit{agent\_sota} denotes the frozen PPO GridNet checkpoint distributed with Gym-$\mu$RTS \citep{huang_gym-rts_2021}. GridNet produces actions for multiple units through a spatial policy, with invalid-action masking restricting the available choices. \textit{RAISocketAI}, the 2023 competition winner, builds on PPO with convolutional spatial policies, multiple value heads, iterative fine-tuning, and transfer learning to particular maps \citep{goodfriend_competition_2024}. Together, these opponents test performance against fixed scripts, online search, and independently trained neural policies. The training versus holdout assignments used in this study are reported in \Cref{tab:model_comparison}.

\textit{TMA} (TMA, Tactical Manager AI) is in the training pool. TMA won the 2024 competition: it was the highest-ranked submission both on the open maps and on the open and hidden maps combined, and only Mayari and CoacAI baselines placed above it in one of the two rankings \citep{mazzaTMA2024,micrortsCompetitionResults2024}. TMA is scripted and does not perform any search. It has four hand-written scripts: a worker or military rush, and a worker or military defense. At every step, a controller chooses one of them through two binary decisions based on features of the game state. It chooses a military script rather than a worker script once it has barracks, when resources accumulate or when the enemy is far from its base. The controller then sets the parameters of the selected script from the map size, unit counts and distances. These parameters are the number of harvesting workers, the resource reserve above which the base trains additional workers, and production priorities. The priorities keep the army balanced across light, heavy and ranged units, favor ranged units when no path to the enemy exists, and shift production toward the enemy's most numerous unit type. TMA can be seen as a hand-written counterpart of the hierarchy studied here, in which a fixed rule chooses a strategy and its parameters and scripted unit behaviors execute it.

\section{Command criteria and issue preconditions}
\label{app:command_criteria}

\subsection{Terminal outcomes}
\label{app:command_adherence}

Command adherence is evaluated independently of game outcome over a separate temporal window for each slot. A window terminates with \emph{success} when the command criterion is satisfied, \emph{failure} when an explicit violation occurs, or \emph{timeout} when its deadline expires without success. For command $c$, the terminal success rate used to adapt its Lagrange multiplier is
\begin{equation}
\widehat p_c=\frac{S_c}{S_c+F_c+T_c},
\label{eq:appendix_terminal_adherence}
\end{equation}
where $S_c$, $F_c$, and $T_c$ count the three outcomes. Progress-based shaping guides learning but does not contribute additional successes.

\Cref{tab:command_outcomes} summarizes the criteria and deadlines. Economy and composition use elapsed ticks; posture and worker policy use only evaluable ticks. Posture requires at least one military unit for OFFENSE or DEFENSE, two for SPLIT, and four for FLANK; worker policy requires at least one worker. Already-satisfied STABILIZE and DOUBLE\_BARRACKS commands remain unmeasured until deterioration makes recovery necessary. Reissuing an unchanged active command preserves its window. Composition windows restart after termination, whereas the other slots require explicit reissue. Consequently, success certifies a single evaluation window rather than sustained compliance throughout the episode.

\begin{table}[htbp]
\centering
\small
\setlength{\tabcolsep}{3pt}
\caption{Command-specific terminal criteria. A dash denotes no explicit failure condition; unmet objectives instead time out at deadline $D$. For composition, deadlines correspond to commands in the listed order.}
\label{tab:command_outcomes}
\begin{tabular}{p{0.22\linewidth}p{0.39\linewidth}p{0.23\linewidth}p{0.09\linewidth}}
\toprule
Command & Success & Failure & $D$ \\
\midrule
STABILIZE & Base, four workers, and a barracks; recovery exceptions below. & Unauthorized second barracks. & 450 \\
GROW & Living worker count increases by at least one from window start. & Unauthorized second barracks. & 300 \\
HOLD & No prohibited worker production for 140 ticks; a base or barracks survives. & New worker production without a worker deficit, or unauthorized second barracks. & 250 \\
DOUBLE\_BARRACKS & A base and at least two barracks survive. & --- & 1200 \\
\midrule
LIGHT, HEAVY, RANGED & At least three new military units; requested type accounts for $\geq75\%$. & At least three new units and $>30\%$ non-target types. & 700, 800, 700 \\
LIGHT\_HEAVY, LIGHT\_RANGED, HEAVY\_RANGED & Both requested types produced; each accounts for $\geq25\%$, together $\geq80\%$. & Both types produced, at least four new units, and $>35\%$ non-target types. & 1300, 1200, 1300 \\
\midrule
OFFENSE & Offensive mode on $\geq50\%$ of evaluable ticks. & --- & 600 \\
DEFENSE & Defensive mode on $\geq60\%$ of evaluable ticks. & --- & 300 \\
SPLIT & Split mode on $\geq40\%$ of evaluable ticks. & --- & 700 \\
FLANK & Flanking formation on $\geq30\%$ of evaluable ticks. & --- & 450 \\
\midrule
ECON\_ONLY & 60 worker-present ticks without a distant-attack violation. & $>5\%$ of workers attack more than six tiles from own structures. & 240 \\
HARASS & A worker occupies the opponent's half for 40 successive worker-present evaluations. & --- & 240 \\
\bottomrule
\end{tabular}
\end{table}

\paragraph{Economy and composition.}
STABILIZE requires only a restored base and three workers if no base existed at window start; an active HARASS window reduces either worker target by one. Unauthorized construction denotes the episode's first transition from one to two barracks without DOUBLE\_BARRACKS. HOLD allows production already in progress and replacement of workers lost relative to the initial count. Composition shares are estimated from accumulated positive changes in living unit counts within the window, excluding the pre-existing army. This proxy can miss production offset by simultaneous losses and can include units whose production began before a command switch. All composition commands additionally time out if no barracks is present at or after tick 350.

\paragraph{Posture and worker policy.}
Military units are classified as offensive or defensive by proximity to observed enemy targets or own bases, barracks, and workers; units approximately equidistant and at least five tiles from both are idle. Let $o,d$ be the offensive and defensive fractions and $a=o+d$. Offensive mode requires $a\geq0.40$, $o\geq0.55$, and $o\geq d$; defensive mode requires $a<0.40$ or $(d\geq0.65\text{ and }d>o)$. Split mode requires $a\geq0.40$ and $o,d\geq0.20$, with neither other mode active. FLANK requires an offensive or defensive core with units on both sides of the own-to-enemy axis: a minority-side unit must have lateral displacement of at least two tiles and normalized forward projection $\geq0.45$ for offense or $\leq0.60$ for defense. All posture successes require at least 50 evaluable ticks, with satisfaction averaged over the window. ECON\_ONLY permits local defense; HARASS measures forward presence rather than attacks, using the opponent's horizontal half on the $16\times16$ map.

\paragraph{Interrupted windows.}
Windows closed by a command change or episode end are not assigned terminal failure or timeout and do not enter \Cref{eq:appendix_terminal_adherence}. A replaced window can still earn interrupted-progress credit in its slot reward (Appendix~\ref{app:reward_structure}).

\subsection{Issue preconditions}
\label{app:issue_preconditions}
ECO\_STABILIZE only when the economy is damaged (no base, fewer than four workers, or no barracks), ECO\_HOLD with at least three workers, ECO\_DOUBLE\_BARRACKS with at most one barracks, WORKER\_HARASS with at least four workers, and POSTURE\_OFFENSE, SPLIT and FLANK with at least one, two and four combat units. All other commands are always issuable. There is no opening exception: at episode start only WORKER\_ECON\_ONLY and POSTURE\_DEFENSE are issuable in their slots. Preconditions gate strategists, never measurement. A violating slot is repaired to the first issuable command of a fixed fallback chain (STABILIZE, HOLD → GROW; DOUBLE\_BARRACKS → HOLD → GROW; HARASS → ECON\_ONLY; FLANK → OFFENSE → DEFENSE; SPLIT, OFFENSE → DEFENSE), and the original order is re-issued as soon as its precondition holds. Re-issue is upgrade-only.

\section{Reward structure}
\label{app:reward_structure}

The executor learns from five reward streams: one for game performance and one for each command slot. The game reward encourages economic and military progress independently of the current command, while the slot rewards provide feedback on its execution. These streams define separate value targets and are combined only through their normalized advantages, with the command-specific multipliers determining the relative priority of adherence. Terminal command outcomes provide the success-rate estimates used to update those multipliers.

\subsection{Game rewards}

The game-performance reward is
\begin{equation}
 r_t^{\mathrm{game}}=10z_t+0.3h_t+0.2b_t+0.6a_t+0.5u_t,
 \label{eq:appendix_game_reward}
\end{equation}
where $z_t$ is the environment's terminal win/loss signal, $h_t$ counts resource-harvest and resource-return actions, $b_t$ counts production orders for bases or barracks, and $u_t$ counts production orders for Light, Heavy, or Ranged units. The attack signal $a_t$ counts attacks directed at enemy-occupied tiles, subtracting attacks directed at friendly-occupied tiles. Worker production carries no direct reward. The terminal signal is $+1$ when the environment declares the controlled player the winner and $-1$ for other environment-declared game endings; it is zero before termination and at a time-limit truncation without an environment-declared outcome.

The intermediate terms are evaluated from issued actions: construction and unit-production rewards are received when production is ordered, and attack rewards do not measure damage or kills. They supply denser feedback than the terminal outcome alone, but need not rank policies identically to win rate. The flat PPO comparison uses the same game reward. 

\subsection{Adherence rewards}

For a command $c$ active in slot $j$, the slot reward combines within-window shaping $h_{t,j}^{c}$, a terminal payment $b_{t,j}^{c}$, and credit for interrupted progress $i_{t,j}$:
\begin{equation}
 r_{t,j}^{\mathrm{slot}}=h_{t,j}^{c}+b_{t,j}^{c}+i_{t,j}.
 \label{eq:appendix_slot_reward}
\end{equation}
Success pays $12$, timeout $-8$, and explicit failure $-6$. Success payments are $4$ for GROW, $24$ for DOUBLE\_BARRACKS, and $6$ for DEFENSE and FLANK. Repeated success or failure payments for the same command in the same slot are suppressed within a 120-tick cooldown; terminal outcomes are still recorded. Timeout penalties are unaffected. All coefficients below refer to the unweighted slot streams used to fit the adherence critic.

\paragraph{Economy.}
STABILIZE receives $0.02$ per tick when the worker count exceeds its window-start value and another $0.02$ when the barracks count exceeds its initial value. A surviving base contributes $0.005$; restoring an initially absent base contributes $0.03$, and retaining at least one barracks when one existed initially contributes $0.01$. GROW receives $0.01$ when the worker count exceeds its initial value. HOLD receives $0.01$ while a base or barracks survives. DOUBLE\_BARRACKS receives $0.03$ when the barracks count exceeds its initial value, or $0.01$ during construction while an initial barracks and a base remain. These terms reward intermediate states within the active window; the terminal criteria of Appendix~\ref{app:command_adherence} determine when the requested recovery, growth, or restraint is achieved.

\paragraph{Composition.}
A single-type command receives $0.03$ per tick once production of its requested type has been observed within the window; a two-type command receives $0.04$ once either requested type has been observed. Active production of a requested type adds $0.04$, and active production of an unrequested military type subtracts $0.04$. Each additional requested unit beyond the first of its type contributes $6$; for two-type commands, this credit requires that both requested types have been observed. Production is estimated from positive changes in living unit counts, as for the composition criteria (Appendix~\ref{app:command_adherence}).

To provide feedback before military production becomes possible, composition also rewards progress toward a barracks. Let $\phi_t$ equal $1$ when a barracks exists, $0.6$ when a worker is constructing a structure without a surviving barracks, and $0.3\min(R_t/5,1)$ otherwise, where $R_t$ is the resource stock. With $m_{t-1}$ the largest milestone value previously reached in the window, the additional reward is $3\max(0,\phi_t-m_{t-1})$. The initial milestone reflects the barracks and resource stock at command issue. Thus, resource accumulation and construction provide bounded intermediate credit, and revisiting a previously reached milestone provides none. Construction of a base can also satisfy the intermediate construction milestone.

\paragraph{Posture and worker policy.}
On evaluable ticks, satisfying the requested posture pays $0.03$ for OFFENSE, $0.015$ for FLANK, and $0.01$ for DEFENSE or SPLIT. When OFFENSE is not satisfied, it instead receives $0.02$ times the offensive fraction of military units. FLANK receives $0.01$ when its lateral-formation condition holds without the required core posture. ECON\_ONLY receives $0.002$ per worker-present tick without a distant-attack violation; HARASS receives $0.002$ when a worker occupies the opponent's half. The spatial predicates and exposure requirements are specified in Appendix~\ref{app:command_adherence}.

\paragraph{Command changes.}
Replacing an unfinished command after at least 30 window ticks contributes $0.3B_cq$ to its slot stream, where $B_c$ is that command's success payment and $q\in[0,1]$ measures progress toward its criterion. This credit preserves a learning signal for partially executed commands when the strategist changes its request. It does not count as terminal success. The shaping terms are task-specific learning signals; we do not assume that they preserve the optimal policy of either the terminal game objective or the terminal command-success objective.

\section{Architecture}
\label{sec:architecture}
A convolutional stem produces 64 full-resolution channels, followed by a stride-two convolution to 128 channels and six conditioned residual blocks. On a $16\times16$ map the bottleneck has spatial size $8\times8$. Group normalization and SiLU activations are used in this architecture. The decoder upsamples to 64 channels and fuses a gated full-resolution skip before predicting per-cell action logits. The critics concatenate global average and maximum pooled features with the command embeddings and use a hidden layer of width 256. Commands have 16-dimensional embeddings concatenated in slot order.

The implementation passes a constant $\boldsymbol\omega=(0.5,0.5)$ through a learned projection. It is fixed throughout constrained training and evaluation, so its contribution is absorbed into the conditioning map $F_\psi$ in the main text; it is not a requested objective trade-off. The flat control uses the constant $(1,0)$.

The actor additionally receives the composition embedding as spatial conditioning. A command-only production head supplies logits for Light, Heavy, and Ranged production, replacing those entries in the spatial actor's unit-type output; logits for workers and structures remain state-conditioned. Single-type commands receive fixed target/non-target logit offsets, while invalid-action masks still apply. The production head receives cross-entropy supervision from composition targets (one-hot for a single type, equal mass over the two types for mixed compositions), with its input embedding detached. Its auxiliary coefficient is $0.10$ until five million learner steps, then decreases linearly to $0.06$ at ten million steps and remains there. The flat control disables this production override and auxiliary loss as well as the adherence objective.

\section{Command sampling for executor training}
\label{app:tuple_bandit}

The generated pool contains 256 random schedules by default, each with one to five phases, uniformly sampled command tuples, and time- or state-based transitions. Coverage sampling balances schedule selections and favors underexposed commands, with a uniform exploration component. Commands with unmet preconditions are temporarily replaced by feasible alternatives and reissued when their preconditions hold. 

The reported run begins with coverage episodes for the first $40\%$ of training, then linearly increases the bandit episode probability to $0.6$ over the next $10\%$. The remaining $0.4$ share continues to use generated schedules. Table learning is active throughout, including the initial coverage period.

\paragraph{Context and evidence.}
The stage is late at tick 1200 or later; before then it is the army stage when at least three combat units are present, the early stage when a barracks exists, and the opening stage when neither condition holds and the tick is below 350. The remaining cases use the early stage. Momentum takes the values behind, even, or ahead by comparing own and observed enemy military counts. 
The evidence key comprises whether enemy barracks have been seen, the set of military unit types seen, whether enemy worker aggression has been observed, a bucket for the maximum observed enemy worker count ($\leq3$, $4$--$6$, or $\geq7$), and a 150-tick time bucket. Opponent identities supply training classes; league snapshots share one class.

The recognizer uses $a_0=0.5$ in \Cref{eq:scouting_belief}. For the weighted sums in \Cref{eq:thompson_score,eq:bandit_mean}, only classes with $b_t^{\mathrm{raw}}(r)>0.01$ are retained and their weights are renormalized. If none exceed the cutoff, the maximal-probability classes are retained instead. If no classes have yet been recorded, selection uses a symmetric Beta prior directly.

\paragraph{Slot-additive prior.}
For a fixed $(r,x)$ row, write $W_{\mathbf{g}},L_{\mathbf{g}}$ for tuple counts and $W_{j,c},L_{j,c}$ for the corresponding slot marginals. The smoothed base rate and command-specific marginal rate are
\begin{equation}
p_0=\frac{1+\sum_{\mathbf{g}}W_{\mathbf{g}}}
{2+\sum_{\mathbf{g}}(W_{\mathbf{g}}+L_{\mathbf{g}})},
\qquad
p_{j,c}=\frac{W_{j,c}+\kappa p_0}{W_{j,c}+L_{j,c}+\kappa}.
\label{eq:slot_marginals}
\end{equation}
The tuple prior used in \Cref{eq:bandit_beta} is
\begin{equation}
m_{r,x,\mathbf{g}}=\sigma\!\left[
\operatorname{logit}(p_0)+\sum_{j=1}^{4}
\left(\operatorname{logit}(p_{j,g_j})-\operatorname{logit}(p_0)\right)
\right],
\label{eq:slot_prior}
\end{equation}
where $\sigma$ is the logistic function and $\kappa=4$. A command without marginal data contributes zero deviation from the base rate. Probabilities are clipped away from zero and one before taking logits.

\paragraph{Outcome credit.}
Accumulate each episode's active duration by context--tuple pair, yielding records $i$ with context $x_i$, tuple $\mathbf{g}_i$, and total duration $d_i$. Repeated visits to the same pair are combined before filtering. Records with fewer than 40 ticks are excluded; if all are shorter, the longest carries the label. For retained records $\mathcal{I}$, set $w_i=d_i/\sum_{\ell\in\mathcal{I}}d_\ell$ and let $y=1$ for a win and $y=0$ otherwise, including a draw. For each touched class--context row, all tuple and slot counts are first multiplied by the discount factor $\xi=0.995$. The tuple update can be written
\begin{equation}
\begin{pmatrix}W_{r,x,\mathbf{g}}\\L_{r,x,\mathbf{g}}\end{pmatrix}
\leftarrow \xi\begin{pmatrix}W_{r,x,\mathbf{g}}\\L_{r,x,\mathbf{g}}\end{pmatrix}
+\sum_{\substack{i\in\mathcal{I}:\,x_i=x,\\\mathbf{g}_i=\mathbf{g}}}
 w_i\begin{pmatrix}y\\1-y\end{pmatrix}.
\label{eq:bandit_credit}
\end{equation}
The same fractional increments are added to each active command's slot marginal. Discounting happens once per touched row per episode. This assigns observational credit to commands used in a game; it does not estimate their causal effect on winning.

\paragraph{Candidate selection and switching.}
The training tolerance in \Cref{eq:bandit_selection} is $\delta=0.05$. At the opening all issuable tuples are candidates. Mid-game candidates normally differ in at most one slot from the current tuple; an empty local candidate set falls back to the full issuable set. Decisions are due every 150 ticks or after a changed barracks/type/aggression evidence signature persists for two fresh feature checks, and a tuple must remain active for at least 60 ticks before it can be replaced. A switch away from a still-feasible current tuple requires a score improvement greater than $0.03$. Evaluation uses posterior means and $\delta=0$, with the saved usage counts breaking exact score ties before random tie-breaking; neither usage nor payoff counts are updated.

\section{Training details}
\label{app:training}

\subsection{Hyperparameters}
\label{sec:hyperparameters}
We train for 240M learner steps using 72 environments across three $16\times16$ maps (Appendix~\ref{sec:curriculum}). Of these, 40 run against scripted bots and 32 form 16 self-play pairs. Each 512-step rollout collects 28,672 transitions from the 56 learner environments. We use 8 minibatches and 2 epochs per update, for approximately 8,370 updates in total.

The learning rate starts at $2\times10^{-4}$ and decreases linearly to a floor of $3\times10^{-5}$, reached at $85\%$ of training. The entropy coefficient decreases from 0.005 to 0.001. We use a PPO clipping threshold of 0.1, a KL early-stopping threshold of 0.02, a discount factor $\gamma=0.99$, and a GAE decay of $0.95$.

\subsection{Multiplier settings}
\label{app:multiplier_settings}
All commands use a target of $\tau_c=0.8$, an initial multiplier of $0.3$, a step size of $\eta_\lambda=0.01$, bounds of $[0.05,1.0]$, and a smoothing coefficient of $\nu=0.9$. A command's multiplier is updated only if the rollout contains at least five terminal events for that command. The first eligible rollout initializes its smoothed rate directly. The stall guard counts consecutive eligible updates in which the rate is below target and the multiplier is at least $90\%$ of its cap. If 100 such updates pass without a rate improvement greater than $0.02$, the multiplier is held at its lower bound. Multiplier updates resume once the rate improves by more than $0.02$ relative to the reference rate recorded when it was held.

\subsection{Curriculum and self-play league}
\label{sec:curriculum}
The learner always controls player~0. At the start of each episode, both scripted and self-play games sample one of the \texttt{basesWorkers16x16} maps A, D and I with probabilities $1/2$, $1/4$ and $1/4$, respectively. The flat comparison uses the same curriculum, maps, league rules and training budget.

\paragraph{Scripted curriculum.} Training follows five phases with lengths in the ratio $1{:}1{:}2{:}3{:}3$. These span updates 1--837, 838--$1\,674$, $1\,675$--$3\,348$, 3$\,349$--$5\,859$ and $5\,860$--$8\,370$, corresponding to approximately 24M, 24M, 48M, 72M and 72M learner steps. The current phase is determined by training progress. At each phase change, the 40 scripted environments are allocated to bots according to the new phase's mixture weights. These assignments are shuffled across environment slots, and each slot keeps its assigned bot until the next phase. \Cref{tab:curriculum} gives the exact number of environments assigned to each bot.

\textit{lightRushAI} and \textit{rojo} are introduced in phase B, \textit{mayari} in C, and \textit{coacAI} and \textit{TMA} in D. In the final phase, \textit{coacAI}, \textit{mayari} and \textit{TMA} occupy 30 of the 40 scripted environments. Training uses no other scripted bots, so the six held-out opponents in \Cref{tab:model_comparison} are never encountered.

\paragraph{Self-play league.} In the 16 self-play pairs the learner plays a frozen policy drawn from a pool that holds the \textit{agent\_sota} checkpoint at full strength and snapshots of the learner taken every 200 updates. At most eight snapshots are kept: when a ninth is added, the snapshot the learner beats most often over at least five recorded games is removed, or the oldest if none has five games; \textit{agent\_sota} is never removed. When a pair's game ends, its next opponent is drawn by prioritized fictitious self-play. With $p$ the learner's win rate against an opponent over their last 32 games, estimated under a uniform Beta prior, each opponent is weighted by $p(1-p)$, which favors opponents of about equal strength, and the normalized weights are mixed with a $10\%$ uniform share. Snapshots of the strategist--executor take their commands from the learner's current payoff table, read-only and by posterior mean, once the learner itself uses the bandit, and from generated schedules before that; snapshots of the flat policy play its constant tuple.

\begin{table}[h]
\centering
\small
\caption{Scripted-bot environments per curriculum phase (out of 40).}
\label{tab:curriculum}
\begin{tabular}{@{}lrrrrr@{}}
\toprule
Opponent & A & B & C & D & E \\
\midrule
randomBiasedAI & 24 & 18 & 8  & 2  & 2 \\
workerRushAI   & 16 & 8  & 4  & 4  & 4 \\
lightRushAI    & -- & 10 & 8  & 4  & 4 \\
rojo           & -- & 4  & 4  & -- & -- \\
mayari         & -- & -- & 16 & 14 & 8 \\
coacAI         & -- & -- & -- & 8  & 14 \\
TMA            & -- & -- & -- & 8  & 8 \\
\midrule
Learner steps  & 24M & 24M & 48M & 72M & 72M \\
\bottomrule
\end{tabular}
\end{table}

\section{Constraint satisfaction and controllability audit}
\label{app:constraints}

\subsection{Constraint satisfaction}
\label{app:final_duals}
\Cref{tab:final_duals} gives the multiplier and smoothed terminal success rate of every command stored with the final checkpoint of the strategist--executor run, which \Cref{sec:controllability} summarizes. The multiplier settings are those of Appendix~\ref{app:multiplier_settings}.

\begin{table}[h]
\centering
\small
\setlength{\tabcolsep}{4pt}
\caption{Multipliers $\lambda_c$ and smoothed success rates $\bar\rho_c$ stored with the final checkpoint (target $0.8$, bounds $[0.05, 1.0]$). No constraint is flagged as stalled.}
\label{tab:final_duals}
\begin{tabular}{lcc@{\hskip 18pt}lcc}
\toprule
Command & $\lambda_c$ & $\bar\rho_c$ & Command & $\lambda_c$ & $\bar\rho_c$ \\
\midrule
\texttt{STABILIZE}         & 0.05 & 0.94 & \texttt{LIGHT\_RANGED} & 0.05 & 0.92 \\
\texttt{GROW}              & 0.20 & 0.82 & \texttt{HEAVY\_RANGED} & 0.05 & 0.91 \\
\texttt{HOLD}              & 0.45 & 0.79 & \texttt{OFFENSE}       & 0.24 & 0.79 \\
\texttt{DOUBLE\_BARRACKS}  & 0.05 & 1.00 & \texttt{DEFENSE}       & 0.05 & 0.96 \\
\texttt{LIGHT}             & 0.83 & 0.80 & \texttt{SPLIT}         & 0.05 & 1.00 \\
\texttt{HEAVY}             & 0.47 & 0.75 & \texttt{FLANK}         & 0.29 & 0.92 \\
\texttt{RANGED}            & 0.49 & 0.80 & \texttt{ECON\_ONLY}    & 0.31 & 0.79 \\
\texttt{LIGHT\_HEAVY}      & 0.05 & 0.90 & \texttt{HARASS}        & 0.06 & 0.79 \\
\bottomrule
\end{tabular}
\end{table}

\subsection{Intervention audit}
\label{app:intervention}
\paragraph{Protocol.} The audit uses the final checkpoint of the strategist--executor run without further training, on the training map with full observation and sampled actions. All games open with \texttt{STABILIZE}, \texttt{LIGHT}, \texttt{DEFENSE}, \texttt{ECON\_ONLY}. The intervention tuple is installed at the first tick in $[250, 900]$ at which the executor owns a base, a barracks, at least four workers and at least one military unit, so every issue precondition holds. Commands are pinned at every step. The six settings are the reference (\texttt{HOLD}, \texttt{LIGHT}, \texttt{DEFENSE}, \texttt{ECON\_ONLY}) and five tuples that replace one slot by \texttt{GROW}, \texttt{HEAVY}, \texttt{RANGED}, \texttt{OFFENSE} or \texttt{HARASS}. Settings within a seed share the Python, NumPy and Torch seeds. For coacAI all 40 reference--intervention pairs had identical observation, mask and action histories up to the intervention and the same game state at it; its remaining randomness makes trajectories diverge afterwards. For droplet and obiBotKenobi only 1 of 36 and 0 of 31 pairs were state-matched, because these bots are stochastic from the first tick, so their comparisons are blocked by seed and not matched by state. Ten games against these two opponents ended or stalled before the gate was reached (four against droplet, six against obiBotKenobi); no command was installed in them and they are excluded from every table, including the win counts. Windows last up to 900 ticks and games are capped at 2000.

\paragraph{Measurement.} Unit births are counted from new unit identifiers assigned by the game, not from changes in unit counts, and include production already in progress at the intervention. Physical metrics use the first 300 ticks after the intervention when both games of a pair last that long and the whole observed window otherwise; games against droplet typically end within 300 ticks because the executor wins. Pairs without military production are excluded from composition shares, which leaves 19 to 23 of 24 pairs per command. Intervals in \Cref{tab:controllability} resample pairs ($10\,000$ draws), are not corrected for multiple comparisons, and do not include training-seed uncertainty: the audit covers one trained executor. \Cref{tab:intervention_contract} gives the contract outcomes of the same games. Contract success for composition is lower against droplet and obiBotKenobi than against coacAI because games end before the 700 to 800-tick composition windows resolve; the produced units match the request in every case.

\begin{figure}[t]
    \centering
    \includegraphics[width=\linewidth]{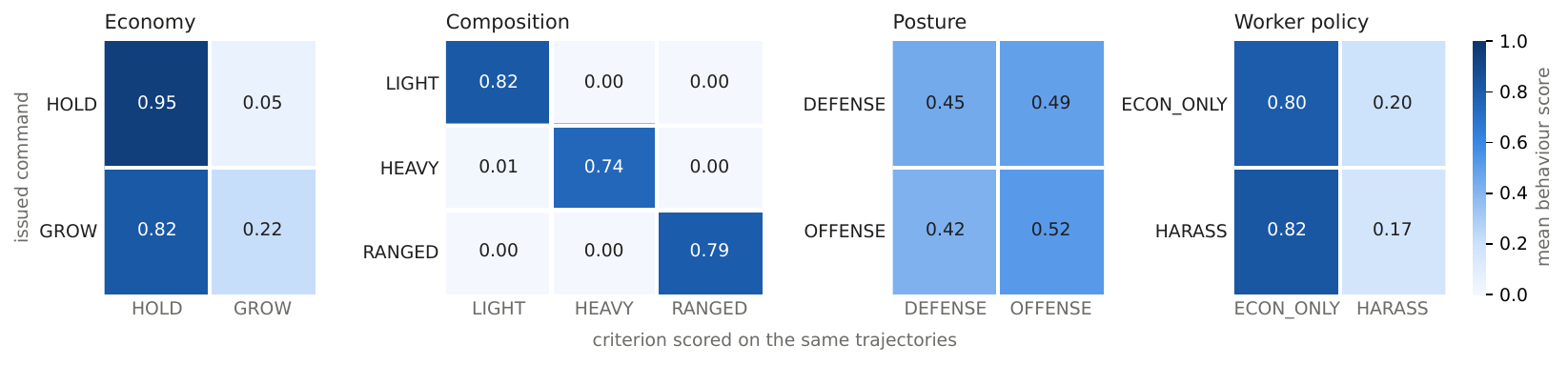}
    \caption{Behavior scores in the intervention games (134 games, three opponents). Rows give the command issued in a slot, columns the compared command whose behavior is scored on the same trajectories. A slot the executor follows is diagonal. Composition is; the rows of posture and worker policy are indistinguishable, so the issued command does not change which behavior is observed. Scores are the heuristic behavior scores of Appendix~\ref{app:intervention}, not contract outcomes.}
    \label{fig:controllability_matrix}
\end{figure}

\paragraph{Behavior scores.} \Cref{fig:controllability_matrix} scores each game against several commands of the varied slot, not only the one that was issued, so it uses heuristic behavior scores rather than contract outcomes. From the intervention until the window closes, every compared command receives a score in $[0,1]$ at each tick, computed from the observed state. For economy, \texttt{GROW} scores worker growth since the intervention and whether a base is training workers, and \texttt{HOLD} the absence of both. For composition, a command scores the requested type's share of the military units produced since the intervention, whether that type was produced at all and whether it is in production, with a penalty while another military type is in production. For posture, \texttt{OFFENSE} scores the fraction of the army beyond $0.45$ of the axis from the mean position of own structures and workers to that of the enemy, and the army's mean position on that axis, and \texttt{DEFENSE} the fraction within five tiles of own structures and the fraction not beyond $0.45$. For worker policy, \texttt{HARASS} scores the fraction of workers in the opponent's half and the fraction attacking, and \texttt{ECON\_ONLY} harvesting activity with workers kept home and not attacking. A game's score for a command is its mean over the ticks at which at least one command of the slot scores above zero; the figure averages over games. These scores agree with the physical metrics of \Cref{tab:controllability}: against the same-seed reference, \texttt{OFFENSE} changes none of the four posture scores and \texttt{HARASS} neither worker-policy score by more than $0.04$ on a $[0,1]$ scale, whereas \texttt{HEAVY} and \texttt{RANGED} raise the score of the requested type by $0.78$ and $0.80$ (\Cref{fig:controllability_matrix}).

\begin{table}[h]
\centering
\small
\caption{Intervention games: windows with at least one contract success for the varied slot, and game outcomes (wins/losses), over the games in which the intervention was installed (at most eight per cell). The reference row has no varied slot.}
\label{tab:intervention_contract}
\begin{tabular}{lcccccc}
\toprule
 & \multicolumn{3}{c}{Contract success} & \multicolumn{3}{c}{Wins/losses} \\
\cmidrule(lr){2-4}\cmidrule(lr){5-7}
Setting & coacAI & droplet & obiBotKenobi & coacAI & droplet & obiBotKenobi \\
\midrule
reference        & --- & --- & --- & 8/0 & 8/0 & 7/0 \\
\texttt{GROW}    & 4/8 & 3/7 & 8/8 & 7/1 & 7/0 & 8/0 \\
\texttt{HEAVY}   & 8/8 & 3/6 & 3/6 & 0/8 & 6/0 & 4/2 \\
\texttt{RANGED}  & 8/8 & 4/7 & 7/8 & 8/0 & 7/0 & 1/7 \\
\texttt{OFFENSE} & 0/8 & 3/8 & 0/7 & 8/0 & 8/0 & 7/0 \\
\texttt{HARASS}  & 6/8 & 6/8 & 3/6 & 8/0 & 8/0 & 6/0 \\
\bottomrule
\end{tabular}
\end{table}

\section{Strength of the opponents on the training maps}
\label{app:opponent_ranking}
The ranks in \Cref{tab:model_comparison} order the opponents by their strength on the training maps. To measure this strength we play every opponent against every other opponent on the three $16\times16$ maps used for training (\texttt{basesWorkers16x16} A, D and I) and fit a Bradley-Terry model to the outcomes. The Bradley-Terry model assigns each player $i$ a scalar strength $s_i$ such that the probability that $i$ beats $j$ is $\sigma(s_i - s_j + h)$, where $\sigma$ is the logistic function and $h$ is a shared seat term for the player that can move first (player~0). A difference of one unit in $s$ corresponds to a win probability of $0.73$; the absolute level is arbitrary and fixed by $\sum_i s_i = 0$. 

The pool contains the twelve opponents of \Cref{tab:model_comparison} and the two other training bots, \textit{rojo} and \textit{randomBiasedAI}. Bots are constructed exactly as in the training environment (same Java classes, unit-type table, time budgets, full observability and the $2\,000$-step cap) and play each other in the same MicroRTS client, so their strength here is the strength the executor faced. Every ordered pair plays on every map, i.e.\ both seats: 10 games per cell, 5 for \textit{RAISocketAI}, and 4 for a bot against itself. \textit{agent\_sota} plays only as player~0, since our harness runs neural policies in that seat, and only on map A: the GridNet checkpoint was trained on map A alone and does not function on the shifted base positions of maps D and I, where it loses every game, including against \textit{randomBiasedAI}. Its strength is therefore only the map-A strength, identified through the shared seat term, and its row in \Cref{tab:model_comparison} should be read the same way. Most bots are deterministic and their random number generators are not seeded, so the replicate games of a cell are often identical; such a cell is collapsed to one game before fitting (191 of the 520 cells), leaving $2\,993$ effective games. A draw (the step cap or simultaneous elimination) counts as half a win for each side. A weak Gaussian prior ($\mathrm{sd}=3$) on $s$ keeps the strength of a bot that never wins or never loses finite, and confidence intervals come from $1\,000$ bootstrap resamples of the effective games. The per-map fits are centered separately, so strengths are comparable within a column, not across maps.

\begin{table}[h]
\centering
\small
\setlength{\tabcolsep}{4pt}
\caption{Bradley-Terry strength of the opponents on the training maps. Pooled is the fit over all three maps with 95\% bootstrap intervals; A, D and I are separate per-map fits; as P1 is the pooled strength of the bot in the player-1 seat, which is the seat it occupies in \Cref{tab:model_comparison} (each bot enters that fit as two players, one per seat, with $h$ fixed to $0$). Win is the raw fraction of games won with draws counted as half. Split is as in \Cref{tab:model_comparison}: train marks opponents in the executor's scripted curriculum or self-play league. \textit{agent\_sota} plays only on map A and only as player~0 (see text), so its pooled strength is its map-A strength.}
\label{tab:opponent_ranking}
\begin{tabular}{llrlrrrrr}
\toprule
Opponent & Split & Pooled $s$ & 95\% CI & A & D & I & as P1 & Win \\
\midrule
RAISocketAI    & holdout  & $3.10$  & $[2.77, 3.56]$   & $4.96$  & $3.06$  & $2.66$  & $2.73$  & $0.91$ \\
TMA            & train   & $2.59$  & $[2.29, 2.94]$   & $2.75$  & $2.83$  & $2.99$  & $3.19$  & $0.87$ \\
agent\_sota (map A) & train   & $2.30$ & $[2.03, 2.66]$ & $2.72$ & --      & --      & --      & $0.83$ \\
obiBotKenobi   & holdout  & $1.85$  & $[1.67, 2.08]$   & $2.26$  & $2.66$  & $1.39$  & $1.86$  & $0.79$ \\
mayari         & train   & $1.16$  & $[0.94, 1.41]$   & $0.69$  & $1.24$  & $2.05$  & $0.71$  & $0.68$ \\
coacAI         & train   & $0.69$  & $[0.45, 0.98]$   & $1.31$  & $-1.05$ & $1.48$  & $1.05$  & $0.56$ \\
droplet        & holdout  & $0.36$  & $[0.20, 0.55]$   & $-0.28$ & $0.69$  & $0.93$  & $0.42$  & $0.56$ \\
aggrobot       & holdout  & $0.16$  & $[-0.07, 0.43]$  & $-0.16$ & $0.61$  & $0.36$  & $0.12$  & $0.56$ \\
workerRushAI   & train   & $-0.09$ & $[-0.33, 0.15]$  & $-0.43$ & $0.64$  & $-0.29$ & $-0.10$ & $0.55$ \\
rojo           & train   & $-1.04$ & $[-1.36, -0.74]$ & $-1.52$ & $-0.17$ & $-1.23$ & $-0.79$ & $0.28$ \\
lightRushAI    & train   & $-1.18$ & $[-1.56, -0.85]$ & $-1.38$ & $-1.44$ & $-0.63$ & $-1.02$ & $0.36$ \\
rangedRushAI   & holdout  & $-1.36$ & $[-1.83, -0.98]$ & $-2.50$ & $-0.39$ & $-1.35$ & $-0.99$ & $0.27$ \\
heavyRushAI    & holdout  & $-2.15$ & $[-2.58, -1.79]$ & $-2.07$ & $-2.27$ & $-2.61$ & $-1.83$ & $0.22$ \\
randomBiasedAI & train   & $-6.40$ & $[-7.36, -5.83]$ & $-6.34$ & $-6.40$ & $-5.76$ & $-6.27$ & $0.01$ \\
\midrule
Seat term $h$  &     & $0.04$  & $[-0.07, 0.15]$  & $-0.25$ & $0.02$  & $0.25$  & $0$     & \\
\bottomrule
\end{tabular}
\end{table}

\Cref{tab:opponent_ranking} shows clear tiers. \textit{RAISocketAI} is the strongest opponent on every map but I. \textit{tma} and \textit{agent\_sota} (on map A) follow and are not separable from each other, then \textit{obiBotKenobi}: the 2023 and 2024 competition winners, the 2023 runner-up and the GridNet checkpoint are the opponents that matter. \textit{mayari} and \textit{coacAI} form the next tier, then \textit{droplet}, \textit{aggrobot} and \textit{workerRushAI}, which are close to each other and to even strength. The remaining rush bots and \textit{rojo} lose to most of the pool, and \textit{randomBiasedAI} wins $1\%$ of its games. The perfect scores in \Cref{tab:model_comparison} against \textit{lightRushAI}, \textit{workerRushAI}, \textit{heavyRushAI} and \textit{rangedRushAI} are therefore against the weak half of the pool, and the differences between flat and bandit appear against opponents from the upper and middle tiers: \textit{agent\_sota}, \textit{obiBotKenobi}, \textit{droplet} and \textit{aggrobot}.

Two properties of the scripted opponents matter for reading \Cref{tab:model_comparison}. First, several bots are seat- and map-dependent. \textit{coacAI} as player~0 loses all ten games to \textit{workerRushAI} on map A and wins all ten as player~1; pooled over seats and maps it ranks sixth, in the player-1 seat it is the fourth-strongest opponent, and on map D it drops to tenth. The seat term itself is close to zero when pooled but has opposite signs on A (player~1 favoured) and I (player~0 favoured). Our executor plays as player~0 on map A, which is the less favoured seat there. Second, no bot has difficulty playing itself: 20 of the 36 mirror cells end in four identical draws at the step cap, the other deterministic mirrors give every game to the same seat, and only \textit{obiBotKenobi}, \textit{droplet} and \textit{RAISocketAI} vary between mirror games. Mirror games enter the fit only through the seat term.

\section{Model comparison: game counts and fixed tuples}
\label{app:model_comparison_details}
\Cref{tab:model_comparison_counts} gives the game counts behind \Cref{tab:model_comparison}. Rates exclude draws: $W/(W+L)$. For each opponent, we compare the Flat and Bandit win/loss counts using a two-sided Fisher exact test, then apply Holm correction across the twelve opponents. The table reports adjusted $p$-values, with significance assessed at $0.05$. The Bandit--Static comparisons use the same test with Holm correction applied separately across their twelve opponents. These tests assess win rates conditional on a decisive game. In the head-to-head row ($^\dagger$), Flat and Bandit play each other, so we instead use an exact two-sided binomial test of Bandit's decisive-game win rate against $0.5$; this separate test is not included in either correction family. The static head-to-head result is from separate games against Flat.

For Static, the executor holds one fixed tuple for the whole game. For each opponent, we select the best tuple on a separate screen of 20 games per tuple, on seeds disjoint from those of the table, among the 18 combinations of \texttt{STABILIZE}, \texttt{GROW} or \texttt{DOUBLE\_BARRACKS} with the six composition commands. The screen picks the highest $W/(W+L)$ and breaks ties by fewer draws, then by shorter games; its games form the non-\texttt{HOLD} cells of Appendix~\ref{app:payoff_matrix}. \texttt{HOLD} is excluded because it halts worker production at about four workers (\Cref{tab:payoff_matrix}). Posture and worker policy stay at \texttt{DEFENSE} and \texttt{ECON\_ONLY}. The selected tuple is then evaluated in 100 fresh games against every opponent.

\begin{table}[h]
\centering
\small
\setlength{\tabcolsep}{4pt}
\renewcommand{\arraystretch}{1.1}
\caption{Wins/losses/draws for \Cref{tab:model_comparison}, in the same opponent ranking order. Static tuples show economy and composition only: \texttt{DOUBLE} abbreviates \texttt{DOUBLE\_BARRACKS}; \texttt{L}, \texttt{H} and \texttt{R} denote Light, Heavy and Ranged. A pair denotes a mixed composition.}
\label{tab:model_comparison_counts}
\begin{tabular*}{\linewidth}{@{\extracolsep{\fill}}rlrrlr@{}}
\toprule
 & & Flat & Bandit & \multicolumn{2}{c}{Static} \\
\cmidrule(lr){5-6}
Rank & Opponent & W/L/D & W/L/D & Selected tuple & W/L/D \\
\midrule
 1 & RAISocketAI         & 1/98/1  & 33/63/4 & \texttt{GROW+L}      & 47/53/0 \\
 2 & TMA                 & 100/0/0 & 95/5/0  & \texttt{DOUBLE+L+R}  & 94/5/1 \\
 3 & agent\_sota (map A) & 60/38/2 & 92/7/1  & \texttt{STABILIZE+L} & 88/10/2 \\
 4 & obiBotKenobi        & 54/45/1 & 97/3/0  & \texttt{GROW+L}      & 100/0/0 \\
 5 & mayari              & 100/0/0 & 98/2/0  & \texttt{STABILIZE+L} & 100/0/0 \\
 6 & coacAI              & 98/2/0  & 100/0/0 & \texttt{GROW+R}      & 100/0/0 \\
 7 & droplet             & 96/4/0  & 96/4/0  & \texttt{GROW+L}      & 95/5/0 \\
 8 & aggrobot            & 99/1/0  & 97/3/0  & \texttt{GROW+L}      & 95/5/0 \\
 9 & workerRushAI        & 97/2/1  & 97/3/0  & \texttt{STABILIZE+L} & 97/3/0 \\
11 & lightRushAI         & 98/2/0  & 100/0/0 & \texttt{DOUBLE+H}    & 99/0/1 \\
12 & rangedRushAI        & 100/0/0 & 100/0/0 & \texttt{GROW+L+R}    & 100/0/0 \\
13 & heavyRushAI         & 100/0/0 & 100/0/0 & \texttt{DOUBLE+L+H}  & 99/1/0 \\
\midrule
--- & Head-to-head       & 3/97/0  & 97/3/0  & \texttt{STABILIZE+L} & 99/1/0 \\
\bottomrule
\end{tabular*}
\end{table}

\section{Playstyle of Flat, Bandit and Static}
\label{app:playstyle}
\begin{figure}[!htb]
    \centering
    \includegraphics[width=\linewidth]{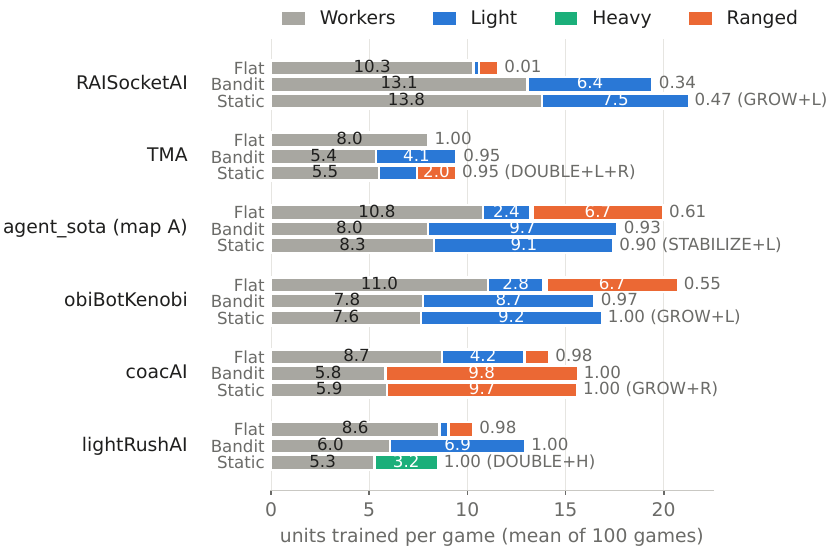}
    \caption{Units trained per game by Flat, Bandit and Static (means over the 100 games of \Cref{tab:model_comparison}, on map A; starting units excluded). The label at each bar gives the win rate $W/(W+L)$ and, for Static, its fixed tuple (\Cref{tab:model_comparison_counts}).}
    \label{fig:playstyle}
\end{figure}

The games of \Cref{tab:model_comparison} record every unit each side trains or loses. \Cref{fig:playstyle} shows what Flat, Bandit and Static train against the four strongest opponents, coacAI and lightRushAI. The bandit plays one build. Against each of these opponents its first barracks is placed at tick 300 in the median game, its army is Light, and it first has two or more military units in the enemy half at a median tick between 550 and 700, or 1030 against RAISocketAI. It also trains fewer workers than Flat against every opponent except RAISocketAI. One exception is coacAI, where the strategist replaces \texttt{LIGHT} with \texttt{RANGED} in all 100 games before the first barracks exists (median tick 242); \texttt{RANGED} scores $1.00$ against coacAI in \Cref{tab:payoff_matrix}, against $0.73$ for \texttt{LIGHT}. Static trains what its fixed tuple names. Where it issues the same composition as the bandit, against agent\_sota, obiBotKenobi and coacAI, the two builds are nearly identical: the same first-barracks tick, worker counts within $0.4$ and army sizes within $0.6$ units per game. The bandit's builds are thus the ones its commands specify.

Flat's build depends on the opponent. Against obiBotKenobi and agent\_sota it places its first barracks at tick 360, trains about three more workers than Bandit and builds a mostly Ranged army. For the command-conditioned executor, \texttt{RANGED} scores $0.11$ and $0.25$ against these two opponents (\Cref{tab:payoff_matrix}), which are also where Flat loses most often after RAISocketAI. Against RAISocketAI, TMA and lightRushAI, Flat trains fewer than two military units per game and never has two in the enemy half in at least 94 of 100 games. This wins against TMA: in 99 of 100 games Flat's workers destroy TMA's barracks at tick 405 and its base at tick 425, and the 100 games contain only six distinct trajectories. Against RAISocketAI the same play loses 98 of 100 games.

\section{Payoff of fixed command tuples}
\label{app:payoff_matrix}
To bound what any strategist could gain from choosing commands per opponent, we play the frozen executor of the Bandit arm under every fixed combination of the four economy and six composition commands, with posture and worker policy at \texttt{DEFENSE} and \texttt{ECON\_ONLY}. Each of the 24 tuples is held for the whole game against the twelve opponents of \Cref{tab:model_comparison} on map A, with 20 games per cell ($5\,760$ games). The 18 tuples without \texttt{HOLD} are the Static screen of Appendix~\ref{app:model_comparison_details}; they are issuable at the opening and run as written. \texttt{HOLD} requires three workers, so the six \texttt{HOLD} tuples are delivered as in training: \texttt{GROW} until three workers exist, then \texttt{HOLD}, which is not reverted. A game scores 1 for a win, 0.5 for a draw and 0 for a loss. We compare the best single tuple on average with an oracle that uses the best tuple for each opponent, both chosen among the 18 tuples that Static selects from. Because the maximum over 18 noisy cells is biased upward, tuples are selected on half of each cell's games and scored on the other half, averaged over $2\,000$ random splits, with ties broken at random.

The choice of commands matters against the strongest opponents. Against RAISocketAI, agent\_sota, obiBotKenobi, coacAI and lightRushAI, the worst tuple scores between $0.00$ and $0.50$ and the best between $0.55$ and $1.00$; against the other seven opponents, every tuple without \texttt{HOLD} scores at least $0.70$. \Cref{tab:payoff_matrix} shows that the best composition also depends on the opponent. \texttt{RANGED} scores $1.00$ against coacAI and $0.11$ against obiBotKenobi, while \texttt{HEAVY} scores $0.00$ against coacAI and above $0.96$ against six other opponents. Nevertheless the oracle gains little. Selected and scored on the same games, which favors the oracle, it scores $0.954$ against $0.921$ for the best single tuple (\texttt{GROW}, \texttt{LIGHT}), a difference of $0.033$, of which $0.021$ comes from coacAI. Out of sample the difference is $-0.011$, with 95\% of splits between $-0.062$ and $+0.029$; including the \texttt{HOLD} tuples gives $-0.015$. The reason is that \texttt{LIGHT} is the best composition, or within $0.02$ of it, against every opponent except coacAI, so knowing the opponent pays only there, by switching to \texttt{RANGED}.

\begin{table}[h]
\centering
\small
\setlength{\tabcolsep}{4pt}
\caption{Mean score of the frozen executor under a fixed composition command, averaged over the economy commands \texttt{STABILIZE}, \texttt{GROW} and \texttt{DOUBLE\_BARRACKS} (60 games per cell), on map A. \texttt{L}, \texttt{H} and \texttt{R} denote Light, Heavy and Ranged; a pair denotes a mixed composition. Ranks are as in \Cref{tab:model_comparison}. The \texttt{HOLD} column averages \texttt{HOLD} over all six compositions (120 games per cell); it is low because \texttt{HOLD} halts worker production: the executor has at most four workers in $95\%$ of these games, against a median peak of six under \texttt{GROW}.}
\label{tab:payoff_matrix}
\begin{tabular}{rlccccccc}
\toprule
Rank & Opponent & \texttt{L} & \texttt{H} & \texttt{R} & \texttt{L+H} & \texttt{L+R} & \texttt{H+R} & \texttt{HOLD} \\
\midrule
 1 & RAISocketAI    & 0.50 & 0.17 & 0.03 & 0.33 & 0.14 & 0.07 & 0.02 \\
 2 & TMA            & 0.95 & 0.97 & 0.97 & 0.94 & 0.95 & 0.95 & 0.28 \\
 3 & agent\_sota    & 0.83 & 0.42 & 0.25 & 0.78 & 0.71 & 0.28 & 0.30 \\
 4 & obiBotKenobi   & 0.97 & 0.58 & 0.11 & 0.82 & 0.77 & 0.36 & 0.38 \\
 5 & mayari         & 0.98 & 0.99 & 1.00 & 1.00 & 1.00 & 1.00 & 0.95 \\
 6 & coacAI         & 0.73 & 0.00 & 1.00 & 0.26 & 0.92 & 0.69 & 0.45 \\
 7 & droplet        & 0.97 & 0.93 & 0.92 & 0.90 & 0.93 & 0.87 & 0.07 \\
 8 & aggrobot       & 0.97 & 0.97 & 0.77 & 0.93 & 0.97 & 0.93 & 0.72 \\
 9 & workerRushAI   & 0.97 & 0.92 & 0.93 & 0.97 & 0.98 & 0.90 & 0.03 \\
11 & lightRushAI    & 1.00 & 0.99 & 0.63 & 1.00 & 0.93 & 0.82 & 0.70 \\
12 & rangedRushAI   & 1.00 & 0.99 & 0.98 & 1.00 & 1.00 & 1.00 & 0.95 \\
13 & heavyRushAI    & 1.00 & 1.00 & 1.00 & 1.00 & 1.00 & 1.00 & 1.00 \\
\midrule
   & Mean           & 0.91 & 0.74 & 0.72 & 0.83 & 0.86 & 0.74 & 0.49 \\
\bottomrule
\end{tabular}
\end{table}




\section{Computation time}
All experiments ran on one server with two NVIDIA H100 NVL GPUs and two AMD EPYC 9534 CPUs (128 cores). Each training run used one GPU for the network and CPU cores for the Java MicroRTS environments. Training to 240M learner steps took approximately 120~h for Flat and 180~h for Bandit, with average throughputs of about 550 and 360 learner transitions per second, respectively. Both runs trained concurrently and shared the machine with other jobs, so these wall-clock times are indicative rather than controlled measurements.

All evaluation games used CPU inference. The model comparison, including the Static screen, took about 3.5~h; the opponent ranking tournament took 1.6~h; the payoff matrix took 0.8~h using 24 parallel processes; and the intervention audit took less than 1~h. These times exclude development runs and discarded variants.

\end{document}